\documentclass[sn-mathphys-num]{sn-jnl}

\usepackage[utf8]{inputenc}
\usepackage[T1]{fontenc}

\usepackage{graphicx}
\usepackage{amsmath,amssymb,amsfonts}
\usepackage{amsthm}
\usepackage{mathtools}
\usepackage{mathrsfs}

\usepackage{booktabs}
\usepackage{multirow}
\usepackage{float}
\usepackage{stfloats}

\usepackage{subcaption}

\usepackage{algorithm}
\usepackage{algorithmic}

\usepackage{listings}

\usepackage[table]{xcolor}

\usepackage{wrapfig}

\usepackage[capitalize,noabbrev]{cleveref}

\usepackage{geometry}
\usepackage{tabularx}

\usepackage{amsthm}

\theoremstyle{thmstyleone}%
\theoremstyle{thmstyletwo}%
\theoremstyle{thmstylethree}%
\usepackage{titletoc}
\usepackage{makecell}

\definecolor{alggray}{gray}{0.45}

\definecolor{bestcol}{RGB}{120,170,245}    
\definecolor{secondcol}{RGB}{170,210,255}  
\definecolor{thirdcol}{RGB}{225,242,255}

\usepackage{algorithm}
\usepackage{algorithmic}
\usepackage{caption}
\usepackage{etoolbox}

\usepackage{algorithm}
\usepackage{algorithmic}
\usepackage{caption}
\usepackage{etoolbox}

\usepackage{graphicx}
\usepackage{multirow}
\usepackage{booktabs}
\usepackage{array}
\usepackage{tikz}
\usetikzlibrary{tikzmark}

\makeatletter
\newenvironment{breakablealgorithm}
  {%
   \begin{center}
   \refstepcounter{algorithm}%
   \hrule height .8pt depth 0pt \kern 2pt
   \renewcommand{\caption}[2][\relax]{%
     {\raggedright \textbf{\ALG@name~\thealgorithm} ##2\par}%
     \ifx\relax##1\relax
       \addcontentsline{loa}{algorithm}{\protect\numberline{\thealgorithm}##2}%
     \else
       \addcontentsline{loa}{algorithm}{\protect\numberline{\thealgorithm}##1}%
     \fi
     \kern 2pt\hrule\kern 2pt
   }%
  }
  {%
   \kern 2pt\hrule\relax
   \end{center}
  }
\makeatother

\begin{document}


\title[CatalyticMLLM]{CatalyticMLLM: A Unified Graph--Text Multimodal Large Language Model for Property Prediction and Inverse Design of Catalytic Materials}


\author[1,2,5]{\fnm{Yanjie} \sur{Li}}\email{liyanjie@semi.ac.cn}
\author*[2,3]{\fnm{Jian} \sur{Xu}}\email{jian.xu@ia.ac.cn}
\author[2,3,6]{\fnm{Xu-Yao} \sur{Zhang}}\email{xyz@nlpr.ia.ac.cn}
\author[2,3]{\fnm{Shiming} \sur{Xiang}}\email{smxiang@nlpr.ia.ac.cn}
\author[4,7]{\fnm{Nian} 
\sur{Ran}}\email{rannian@mail.sic.ac.cn}
\author*[1,2,5]{\fnm{Weijun} \sur{Li}}\email{wjli@semi.ac.cn}
\author*[2,3,6]{\fnm{Cheng-Lin} \sur{Liu}}\email{liucl@nlpr.ia.ac.cn}

\affil[1]{\orgdiv{AnnLab}, \orgname{Institute of Semiconductors, Chinese Academy of Sciences}, \orgaddress{\city{Beijing}, \country{China}}}
\affil[2]{\orgdiv{Zhongguancun Academy}, \orgaddress{\city{Beijing}, \country{China}}}
\affil[3]{\orgdiv{State Key Laboratory of Multimodal Artificial Intelligence Systems}, \orgname{Institute of Automation, Chinese Academy of Sciences}, \orgaddress{\city{Beijing}, \country{China}}}

\affil[4]{\orgdiv{State Key Laboratory of High Performance Ceramics}, \orgname{Shanghai Institute of Ceramics,Chinese Academy of Sciences}, \orgaddress{\city{Shanghai}, \country{China}}}

\affil[5]{\orgdiv{School of Electronic, Electrical and Communication Engineering}, \orgname{University of Chinese Academy of Sciences}, \orgaddress{\city{Beijing}, \country{China}}}

\affil[6]{\orgdiv{School of Artificial Intelligence, University of Chinese Academy of Sciences}, \orgaddress{\city{Beijing}, \country{China}}}

\affil[7]{\orgdiv{Center of Materials Science and Optoelectronics Engineering}, \orgname{University of Chinese Academy of Sciences}, \orgaddress{\city{Beijing}, \country{China}}}

\abstract{
Property prediction and inverse structural design of catalytic materials are typically modeled as two independent tasks: the former predicts target properties from given structures, whereas the latter generates candidate structures conditioned on desired properties. Although this decoupled paradigm facilitates the construction of a ``generation--evaluation--screening'' workflow, the two models may have different domains of applicability because they are trained on different data distributions. Consequently, during closed-loop optimization, the generative model may produce structures that fall outside the reliable applicability domain of the evaluation model, while prediction errors from the evaluator may accumulate over successive iterations, ultimately degrading inverse-design performance.

In this work, we propose CatalyticMLLM, a unified graph--text multimodal large language model for catalytic materials that integrates \textbf{property prediction} and \textbf{inverse design} within a single model and a shared representation space. Under this unified framework, CatalyticMLLM not only performs reliable property prediction by jointly leveraging three-dimensional structures and textual information, but also generates and screens physically feasible CIF candidates conditioned on target properties, thereby establishing a closed-loop workflow of ``inverse design--prediction--screening--redesign.'' Experimental results demonstrate that the unified paradigm outperforms decoupled baselines in both catalytic relaxation-energy prediction and inverse-design tasks, validating the effectiveness of jointly modeling property prediction and structure generation within a single multimodal model.
}

\keywords{Multimodal Large Language Models, Property Prediction, adsorption energy Prediction, Inverse design of materials}

\maketitle
\section{Introduction}
\label{sec:introduction}

The discovery of catalytic materials requires identifying a small number of candidate systems that satisfy the requirements of activity, stability, and selectivity from an enormous space of chemical compositions, crystal structures, and adsorption configurations. For heterogeneous catalysis, material composition, crystal facet orientation, adsorption sites, adsorption configurations, and reaction intermediates are intricately coupled, jointly determining adsorption stability, reaction pathways, and catalytic performance. Adsorption energies therefore, serve as key physical quantities that connect atomic structures with catalytic behavior and also constitute core descriptors for high-throughput screening and catalyst design. Traditionally, structural evaluation of catalytic materials has mainly relied on high-accuracy first-principles methods such as density functional theory. However, such methods are computationally expensive, which limits their high-throughput screening capability over broad chemical spaces and also restricts their scalability to large length and time scales.

In recent years, machine-learning interatomic potentials and graph neural network models have significantly accelerated the rate of materials property prediction. Three-dimensional graph neural networks and equivariant models, such as SchNet\citep{schutt2018schnet}, PaiNN\citep{schutt2021painn}, and Equiformer\citep{liao2023equiformer,liao2024equiformerv2},and so on \citep{klicpera2020dimenetpp,gasteiger2021gemnet,gasteiger2022gemnetoc,batzner2022nequip,batatia2022mace,musaelian2023allegro} have made important progress in energy and force prediction tasks by explicitly modeling local atomic environments, geometric relationships, and rotational symmetries~. These models can effectively capture local interactions in three-dimensional atomic structures, but they are usually designed as specialized predictors that map given structures to target properties. They are difficult to directly apply to natural-language descriptions of catalytic systems, target-property instructions, and design constraints, and they are also not well suited to incorporating expert knowledge through human--model interaction or performing inverse structural generation. Meanwhile, large language models  have provided a new interactive modeling paradigm for materials science. Models based on chemical formulas, SMILES, MOF identifiers, crystal descriptions, and scientific literature have shown that language and symbolic representations can encode valuable chemical knowledge~\citep{goodall2020roost,wang2021crabnet,gupta2022matscibert,matbert2021,moformer2023,transpolymer2023}. For catalytic adsorption-energy prediction, CatBERTa uses natural-language descriptions to regress adsorption-configuration energies, while GAP-CatBERTa further transfers three-dimensional structural knowledge into the textual representation space through graph-assisted contrastive pretraining, thereby improving text-driven prediction performance~\citep{ock2023catalyst,ock2024gapcatberta}. Structure-aware multimodal large language models such as MatterChat~\citep{tang2026multimodal} further demonstrate the potential of aligning materials structure encoders with large language models, enabling models to jointly process materials structures and associated textual prompts and to perform various property prediction and question-answering tasks. However, MatterChat is still mainly oriented toward understanding, question-answering, and property-prediction tasks, and has not yet unified catalytic structure generation, property evaluation, and closed-loop optimization within a single model.

\begin{figure*}[t]
\hspace{-0.4cm}
\centering
\includegraphics[width=134mm]{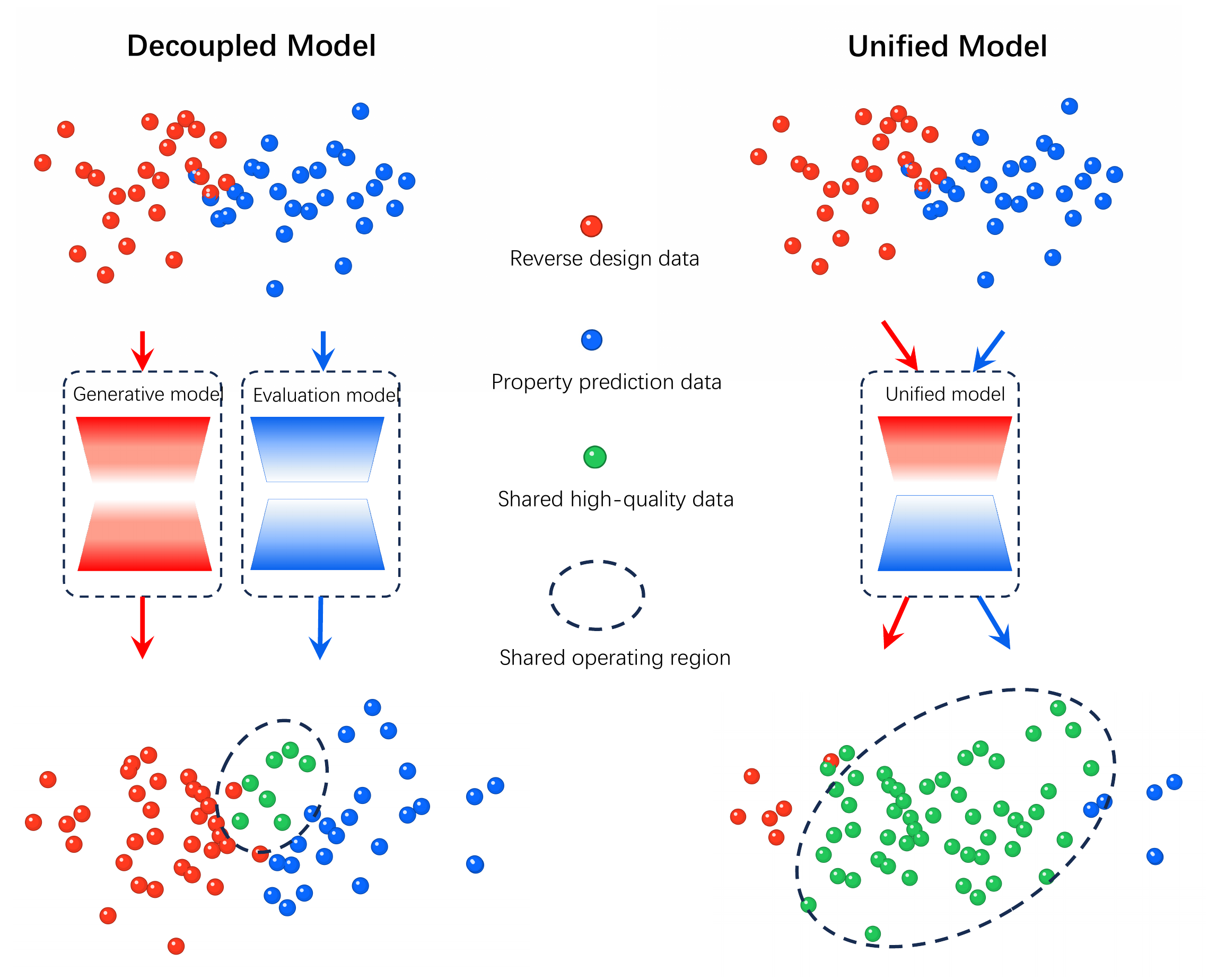}
\caption{
Comparison of the shared operating regions of the decoupled and unified paradigms.
The figure compares two modeling paradigms. In the decoupled paradigm, a generative model and an evaluation model are trained separately to perform inverse design and property prediction, respectively. In the unified paradigm, the two tasks are jointly trained within a single model. When a distribution shift exists between the two training datasets, the unified paradigm facilitates information sharing and synergistic learning across the tasks, thereby yielding a substantially larger shared operating region. By contrast, the effective operating regions of the two independently trained models overlap only marginally in the decoupled paradigm, which may significantly compromise the stability and effectiveness of a closed-loop inverse design system.
}
\label{fig-unified}
\end{figure*}

Using the aforementioned machine-learning models as evaluators, a variety of generative approaches have been developed for crystal structure generation. These include diffusion-based methods, such as CatDRX~\citep{CatDRX}, MatterGen~\citep{zeni2025mattergen}, and related models~\citep{xie2022cdvae,jiao2024spacegroup,yang2023unimat,jiao2023diffcsp}, where, Chemeleon2~\citep{park2026guiding} further enhanced latent diffusion-based crystal generation through reinforcement learning with verifiable multi-objective rewards; flow-based methods, such as CrystalFlow~\citep{luo2025crystalflow}; and autoregressive language models, such as CrystaLLM~\citep{antunes2024crystallm} and other autoregressive approaches~\citep{gruver2024llm,mohanty2025crystext}. However, existing inverse design methods typically still follow a decoupled ``generation--evaluation--screening'' pipeline: a generative model proposes candidate structures, an independent property prediction model or external evaluator scores these candidates, and the candidates are then screened and iteratively refined according to the scores. Because the generator and evaluator may differ substantially in their representation spaces, training objectives, and data distributions, their shared effective operating region can be highly limited as shown in \cref{fig-unified}. Moreover, the evaluator’s output is not only an approximation of the true physical objective but also inevitably reflects its own inductive biases. When the generative model is repeatedly optimized against this evaluator, the search process may gradually become biased toward local structural regions familiar to the evaluator rather than the true physical energy landscape, leading to error accumulation, reduced structural diversity, and diminished stability and reliability of closed-loop optimization.

Recent works such as UniMate and UniGenX~\citep{zhan2025unimate,zhang2025unigenx} have further attempted to move beyond the traditional single-task paradigm by incorporating structure generation, property-related objectives, and conditional constraints into unified modeling frameworks. However, these methods are still mainly centered on diffusion-based or diffusion-hybrid generation, and their optimization processes usually focus on one-shot conditional generation. They have not yet established a rigorous ``generation--prediction--screening--redesign'' closed-loop iterative optimization mechanism for catalytic systems. Moreover, UniGenX only performs sequential encoding of three-dimensional molecular structures, without explicitly introducing an independent three-dimensional geometric encoder or an equivariant structural feature extraction module. At the same time, these methods do not use a natural-language large language model as a unified backbone for semantic interaction, expert-knowledge injection, and CIF-level structure generation. Therefore, there remains considerable room for further development in handling textual catalytic design objectives, contextual modeling, and local redesign of candidate structures, and interactive expression of design constraints.

To address the above issues, this work proposes CatalyticMLLM, a graph--text multimodal large language model for catalytic materials property prediction and inverse design. Unlike conventional decoupled `generation--evaluation--screening'' pipelines, CatalyticMLLM unifies adsorption energy prediction, CIF-level structure generation, candidate screening, and target-driven redesign within a single model framework. Compared with models such as UniMate and UniGenX, CatalyticMLLM further introduces rigorous multimodal fusion training and a closed-loop iterative optimization mechanism, emphasizing the synergy among textual sequences, three-dimensional structures, and the pretrained knowledge of large language models. Our core idea is that property prediction and structure generation should not be regarded as two independent tasks connected by external interfaces, nor should they merely remain as parallel tasks under a shared model framework. Instead, they should be modeled as two complementary directions of the same structure--property distribution, enabling joint modeling and cyclic optimization in a shared representation space. Under this unified framework, CatalyticMLLM is no longer limited to a single property prediction or structure generation task, but serves as a multifunctional unified model for catalytic materials: it can predict adsorption energies by jointly leveraging three-dimensional atomic structures and textual descriptions, generate candidate CIF structures according to target properties and system constraints, and further perform local optimization using current high-quality structures as context when the candidate results do not yet satisfy the target requirements. In addition, CatalyticMLLM forms an `inverse design--prediction--screening--redesign'' closed-loop workflow driven by the same model, enabling generation, prediction, screening, and redesign to be performed collaboratively within a unified representation space, thereby reducing representation shifts, distribution inconsistencies, and evaluation biases introduced by external evaluators. Overall, CatalyticMLLM provides a new multimodal large language model paradigm that integrates property prediction and inverse design, offering a unified framework for structural understanding, property evaluation, candidate generation, and closed-loop optimization of catalytic materials.

\section{Results}

\label{sec:experiments}
This work conducts experiments mainly from two perspectives: the accuracy of property prediction and the precision of inverse design. For property prediction accuracy, we compare our method with language-model-based property prediction approaches such as CatBERTa and GAP-CatBERTa, as well as machine-learning methods including GemNet-OC, EquiformerV2, and UMA\cite{wood2026family}. For inverse design, we select several strong baseline algorithms developed in recent years, including DiffCSP, MAGECS, and so on, for comparison. In addition, we further provide strong experimental evidence demonstrating the advantages of the unified architecture over the separate architecture. Before presenting the experiments, we first provide a brief overview of the overall architecture of CatalyticMLLM.

\subsection{Overview of CatalyticMLLM}
\begin{figure*}[t]
\centering
\includegraphics[width=132mm]{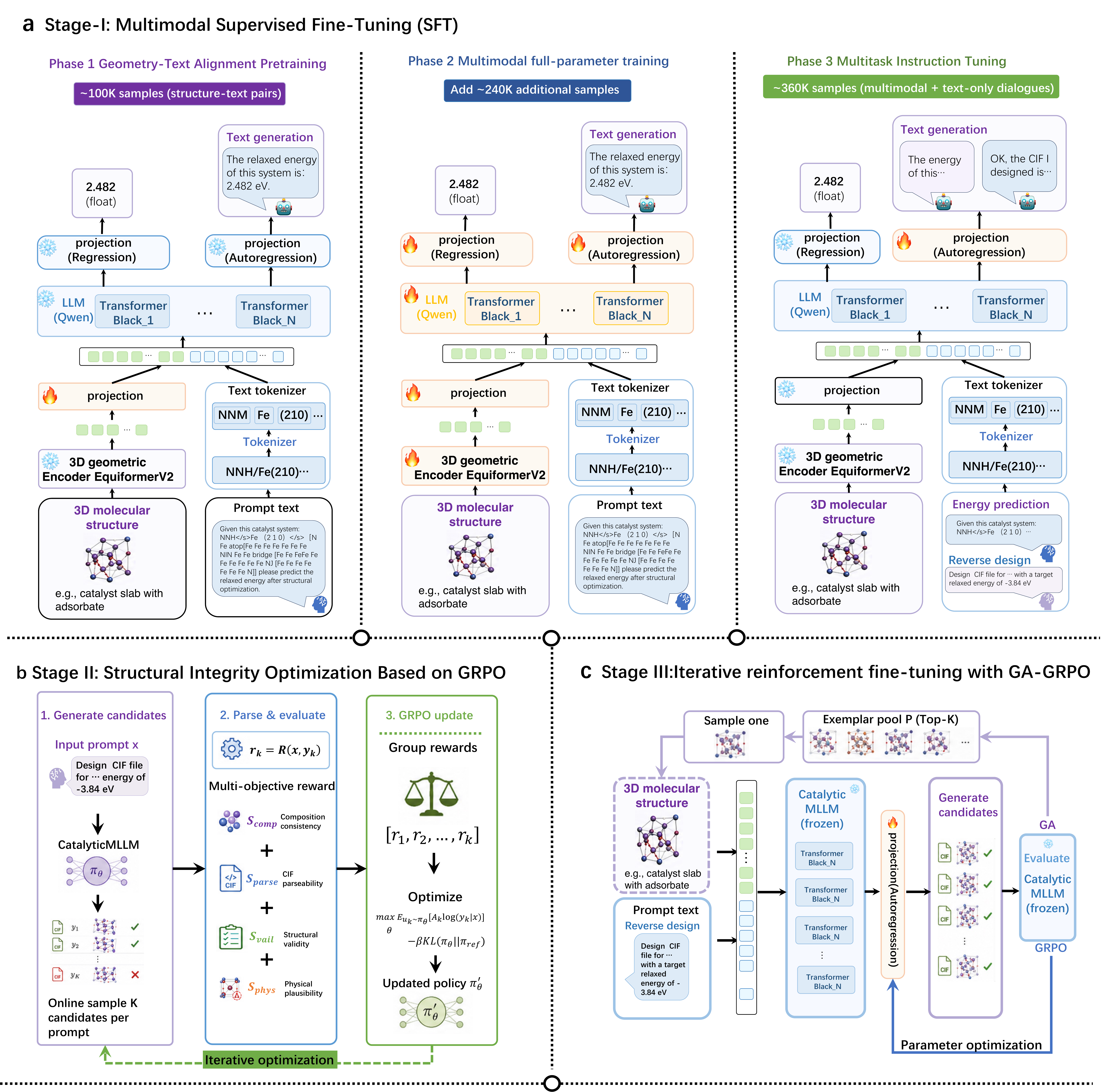}
\caption{
Workflow of the CatalyticMLLM framework. First, large-scale data are used for pre-training, enabling the model to perform property prediction and acquire preliminary inverse design capabilities. Second, GRPO is applied for reinforcement fine-tuning, allowing the model to generate higher-quality CIFs. Third, GA-GRPO is used for further fine-tuning, enabling the model to iteratively optimize the generated CIFs so that they better satisfy the desired target objectives.
}
\label{fig-1}
\end{figure*}
CatalyticMLLM integrates three-dimensional structure encoding, textual semantic modeling, energy prediction, and CIF-level structure generation into a single multimodal large language model. This enables the model to perform forward property prediction, inverse structure generation, candidate evaluation, structure screening, and target-driven redesign within a unified parameter space and a unified latent representation space. This design reduces evaluation bias and error accumulation caused by inconsistencies in representation spaces, training objectives, and data distributions between external evaluators and generators, while also overcoming the limitations of existing diffusion-based generative frameworks in natural-language conditional control, expert-knowledge interaction, and closed-loop iterative optimization. The training data for CatalyticMLLM consist of two types. The first type comprises multimodal data for property prediction, where each input includes a three-dimensional molecular structure and a textual prompt, and the corresponding output is a textual prediction of the target property. The second type consists of text-only data for inverse design, where the input is an inverse-design prompt and the output is a textual response containing the target CIF structure.

Specifically, CatalyticMLLM employs EquiformerV2 as the three-dimensional geometric encoder and Qwen-2.5-VL as the backbone for semantic reasoning, task understanding, and CIF structure generation. A trainable cross-modal projection module is introduced to align atomic geometric representations with the representation space of the language model, thereby enabling three-dimensional geometric perception, textual semantic understanding, property evaluation, and structure generation within a unified framework. The overall architecture and training process of CatalyticMLLM are illustrated in Fig.~\ref{fig-1}. The training process primarily consists of three stages. 

\textbf{Multimodal Supervised Fine-Tuning (SFT).}
In this stage, the model first uses EquiformerV2 to extract geometric representations from the atomic structures of catalytic systems, and then aligns them with the language representation space of Qwen-2.5-VL\citep{qwen25} through a Mapping projection module. Meanwhile, descriptions of catalytic systems, task instructions, and target-property constraints are encoded as textual inputs, enabling the model to acquire preliminary capabilities for adsorption energy prediction and CIF structure generation.

\textbf{Structural Integrity Optimization Based on GRPO.}
To improve the usability and physical reliability of generated CIF structures, we design a PVCP reward function for Group Relative Policy Optimization(GRPO)\citep{shao2024deepseekmath} that constrains the generated outputs from four aspects: CIF parseability, structural validity, compositional consistency, and physical plausibility. This reward function guides the model to actively avoid formatting errors, compositional mismatches, and nonphysical atomic arrangements during generation.

\textbf{Iterative Reinforcement Fine-Tuning with GA-GRPO.}
We further propose Genetic Algorithm-enhanced Group Relative Policy Optimization (GA-GRPO), which combines group relative policy optimization with genetic-algorithm-style candidate updating. In this way, candidate sampling within the same round simultaneously supports policy learning and structural selection, allowing the model to progressively approach the target property while maintaining structural feasibility.

Through these three stages, CatalyticMLLM realizes three-dimensional geometric perception, textual semantic understanding, property prediction, CIF generation, and closed-loop redesign within a single model.

\subsection{property prediction accuracy comparison with baselines}
\subsubsection{Property prediction accuracy comparison with Language-model-based baselines}
\begin{figure*}[htp]
\centering
\hspace{-0.318cm}
\includegraphics[width=132mm]{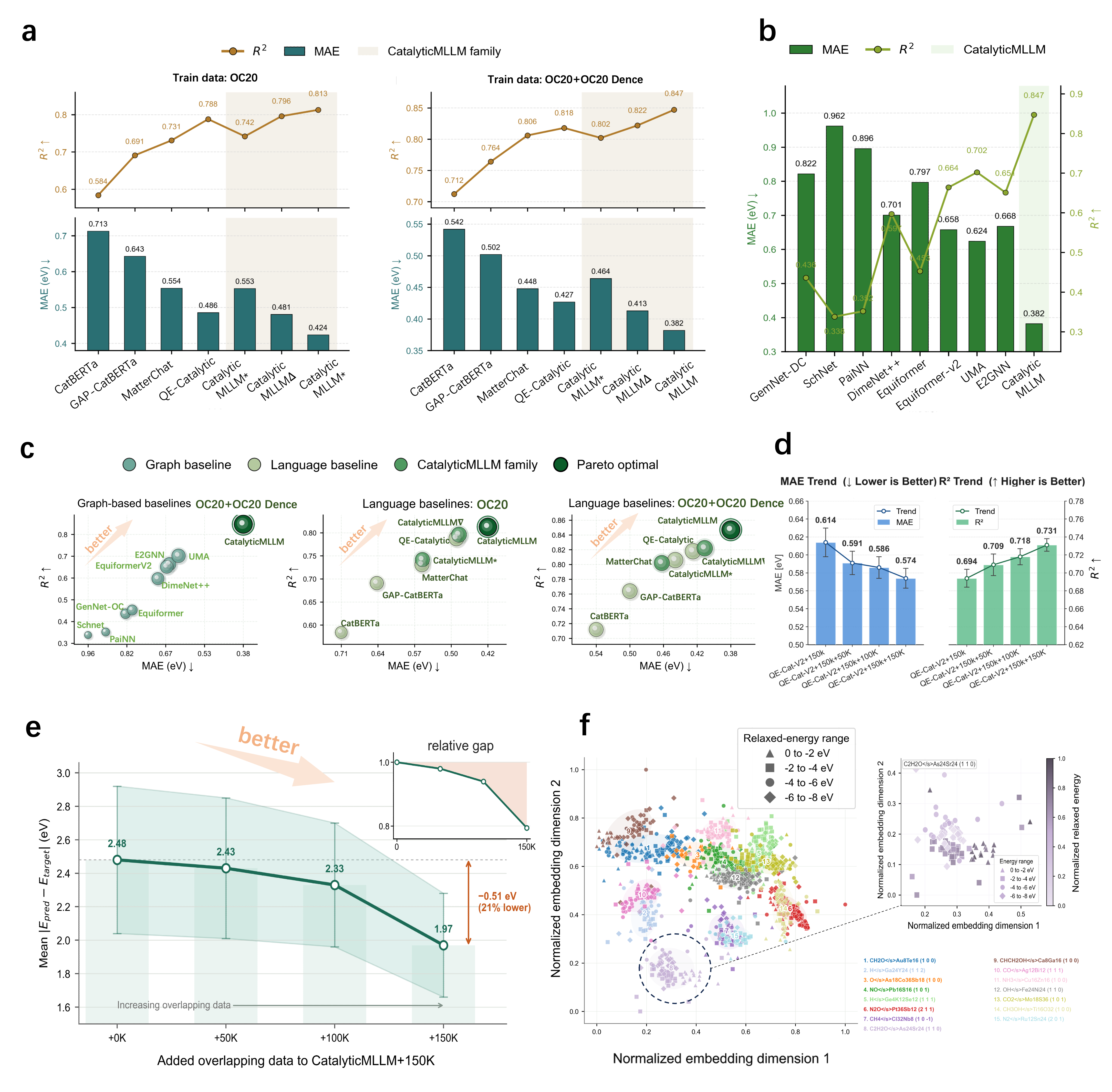}
\caption{
Performance analysis of CatalyticMLLM in property prediction. (a) shows the performance of language-model-based property prediction methods, with training data from OC20 and OC20+OC20-Dense, respectively. (b) compares the performance of conventional graph-based models with CatalyticMLLM. (c) presents the Pareto plots of property prediction performance for language-model-based methods trained on OC20 and OC20+OC20-Dense, respectively, as well as the Pareto plot for conventional graph-based models. (d) and (e) demonstrate that property prediction and inverse design mutually reinforce each other within the unified architecture.(f) visualizes the positions of 3D molecular structural features from different configurations across 15 catalytic systems in the feature space. It can be observed that molecular features belonging to the same catalytic system tend to cluster together in the feature space. Moreover, within the same system, molecular configurations with similar energies tend to be located closer to each other. 
}
\label{fig_1}
\vspace{-0.4cm}
\end{figure*}

We first evaluate CatalyticMLLM against recent language-model-based methods for adsorption energy prediction, including the original CatBERTa and GAP-CatBERTa, the latter of which introduces Graph-Assisted Pretraining (GAP). For a fair comparison, we use identical pretraining and fine-tuning data settings and report mean absolute error (MAE) and coefficient of determination ($R^2$) on the OC20~\citep{chanussot2021open}
and OC20-Dense~\citep{lan2023adsorbml} benchmarks. The results are summarized in~\cref{fig_1}(a).
The MAE and $R^2$ are defined as:

\refstepcounter{equation}\label{eq:mae}
\edef\MAEeqnum{\theequation}

\refstepcounter{equation}\label{eq:r2}
\edef\Rtwoeqnum{\theequation}

\begin{center}
\begin{tabular*}{\linewidth}
{@{\extracolsep{\fill}} c r c r}
$\displaystyle
\mathrm{MAE}
=
\frac{1}{n}\sum_{i=1}^{n}
\left|y_i-\hat{y}_i\right|
$
&
(\MAEeqnum)
&
$\displaystyle
R^2
=
1-
\frac{
\sum_{i=1}^{n}\left(y_i-\hat{y}_i\right)^2
}{
\sum_{i=1}^{n}\left(y_i-\bar{y}\right)^2
}
$
&
(\Rtwoeqnum)
\end{tabular*}
\end{center}

where $n$ denotes the number of samples, $y_i$ and $\hat{y}_i$ represent the ground-truth and predicted values of the $i$-th sample, respectively, and $\bar{y}$ denotes the mean of the ground-truth values.
GAP-CatBERTa consistently outperforms CatBERTa, indicating that incorporating three-dimensional geometric information into text-based pretraining improves the model's ability to distinguish subtle differences among adsorption configurations.

Under the same training data scale, namely 340k OC20 data, CatalyticMLLM* (multimodal training with text-only inference), CatalyticMLLM$^\Delta$ (multimodal training with graph-only inference), and CatalyticMLLM (multimodal inference) all substantially outperform the above text-based baselines. On OC20+OC20 Dense, CatalyticMLLM* achieves an MAE of $0.464$\,eV and an $R^2$ of $0.802$, while the full CatalyticMLLM model further reduces the MAE to $0.382$\,eV and increases $R^2$ to $0.847$. Compared with CatBERTa, this corresponds to an approximately 12.6\% reduction in prediction error and a 19.0\% improvement in predictive performance. More intuitively, Pareto graphs 1 and 2 in \cref{fig_1}(c) clearly show the advantage of CatalyticMLLM (both CatalyticMLLM are in the upper right corner).

Overall, these results show that multimodal training effectively transfers geometric information into the language representation space, allowing text-only inference to exceed strong language-model baselines. In addition, jointly using structural and textual features at inference time further improves prediction accuracy. These findings confirm the importance of both 3D molecular structures and textual descriptions of catalytic systems for property prediction, and demonstrate the strong cross-modal feature fusion capability of CatalyticMLLM.

\subsubsection{Property prediction accuracy comparison with machine learning baselines}

We further compare CatalyticMLLM with a suite of classical atomistic graph neural network baselines, including SchNet, PaiNN, as well as the equivariant-Transformer-based Equiformer, EquiformerV2, and UMA... All 

models are trained on OC20 + OC20-Dense and evaluated on the same test split for adsorption energy prediction; the results are reported in \cref{fig_1}(b).
As shown in \cref{fig_1}(b), among these conventional GNNs, UMA achieves the best performance, outperforming earlier methods. Without changing the training data, CatalyticMLLM reduces the MAE to $0.382\pm0.011$\,eV and improves $R^2$ to $0.847\pm0.008$, indicating that introducing multimodal modeling and alignment yields substantial gains. 
In summary, CatalyticMLLM not only consistently outperforms existing language-model baselines under the ``text+3D'' paradigm, validating the effectiveness of deeply embedding an equivariant geometric encoder into a multimodal large language model, but also exhibits strong advantages in the ``text-only'' and ``graph-only'' setting. The Pareto Figure 3 in ~\cref{fig_1}(c) presents this trend more intuitively.

\subsection{Bidirectional Reinforcement between Property Prediction and Inverse Design in CatalyticMLLM}
\label{sec:mutual_reinforcement}

To investigate whether property prediction and inverse design can mutually reinforce each other within the unified architecture, we conduct two sets of controlled experiments. Specifically, we examine the effect of inverse design data on property prediction and the effect of property prediction data on inverse design. CatalyticMLLM jointly learns the forward mapping from structures to properties and the inverse generation process from target properties to structures within a shared parameter and representation space, enabling knowledge transfer between the two tasks.

\subsubsection{Inverse Design Data Facilitates Property Prediction}

We divide 300k samples into two non-overlapping subsets, $A$ and $B$, each containing 150k samples. The model is first trained on the 150k property prediction samples in subset $A$. While keeping this training set fixed, we progressively add 50k, 100k, and 150k inverse design samples from subset $B$.

As shown in \cref{fig_1}(d), when only the property prediction data are used, the model achieves an MAE of 0.614 eV and an $R^2$ of 0.694. As the amount of inverse design data increases to 50k, 100k, and 150k, the MAE decreases to 0.591 eV, 0.586 eV, and 0.574 eV, respectively, while $R^2$ increases to 0.709, 0.718, and 0.731. These results demonstrate that introducing inverse design data consistently improves property prediction performance when the amount of property prediction data is fixed.

This improvement arises because inverse design requires the model to infer plausible structures from target energies, thereby strengthening its understanding of the correspondence between target energies and local structural patterns. Such inverse constraints improve the modeling of structure--property relationships in the shared representation space and, in turn, benefit forward property prediction.

\subsubsection{Property Prediction Data Facilitates Inverse Design}

We further fix 150k inverse design samples as the base training set and progressively add 0, 50k, 100k, and 150k property prediction samples. After Stage-2 reinforcement fine-tuning, all models are evaluated on the same inverse design test set.

For each target sample, the model generates 10 candidate CIF files. The candidate with the smallest absolute difference between its predicted energy and the target energy is selected as the best result. We then compute the mean optimal energy deviation over the test set:
\begin{equation}
\mathrm{Mean}\left|E_{\mathrm{pred}}-E_{\mathrm{target}}\right|.
\end{equation} 

As shown in \cref{fig_1}(e), when the amount of property prediction data increases from 0 to 50k, 100k, and 150k, the mean deviation decreases from 2.48 eV to 2.43 eV, 2.33 eV, and 1.97 eV, respectively. This result indicates that, with the amount of inverse design data fixed, additional property prediction data improve the alignment between the generated structures and the target properties.

Property prediction training strengthens the model's discriminative understanding of structure--energy relationships, enabling it to identify structural patterns associated with the target energy more accurately during generation. It therefore provides more reliable energy-aware representations and guides the search toward structurally relevant regions.

Overall, property prediction and inverse design represent two complementary directions of the same structure--property distribution: property prediction learns $p(E|S)$, whereas inverse design learns $p(S|E)$. By jointly modeling these two directions through shared parameters and representations, the unified architecture enables bidirectional knowledge transfer. As shown in \cref{fig_1}(f), molecular features from the same catalytic system tend to cluster together in the feature space, while configurations with similar energies are located closer to one another within each system. These observations indicate that CatalyticMLLM learns a well-structured representation of structure--property relationships. The experimental results further confirm that joint training not only improves the two tasks individually but also allows them to reinforce each other, resulting in a more stable and effective closed-loop framework for catalytic materials design.

\begin{figure*}[htp]
\centering
\vspace{-0.8cm}
\hspace{-0.618cm}
\includegraphics[width=134mm]{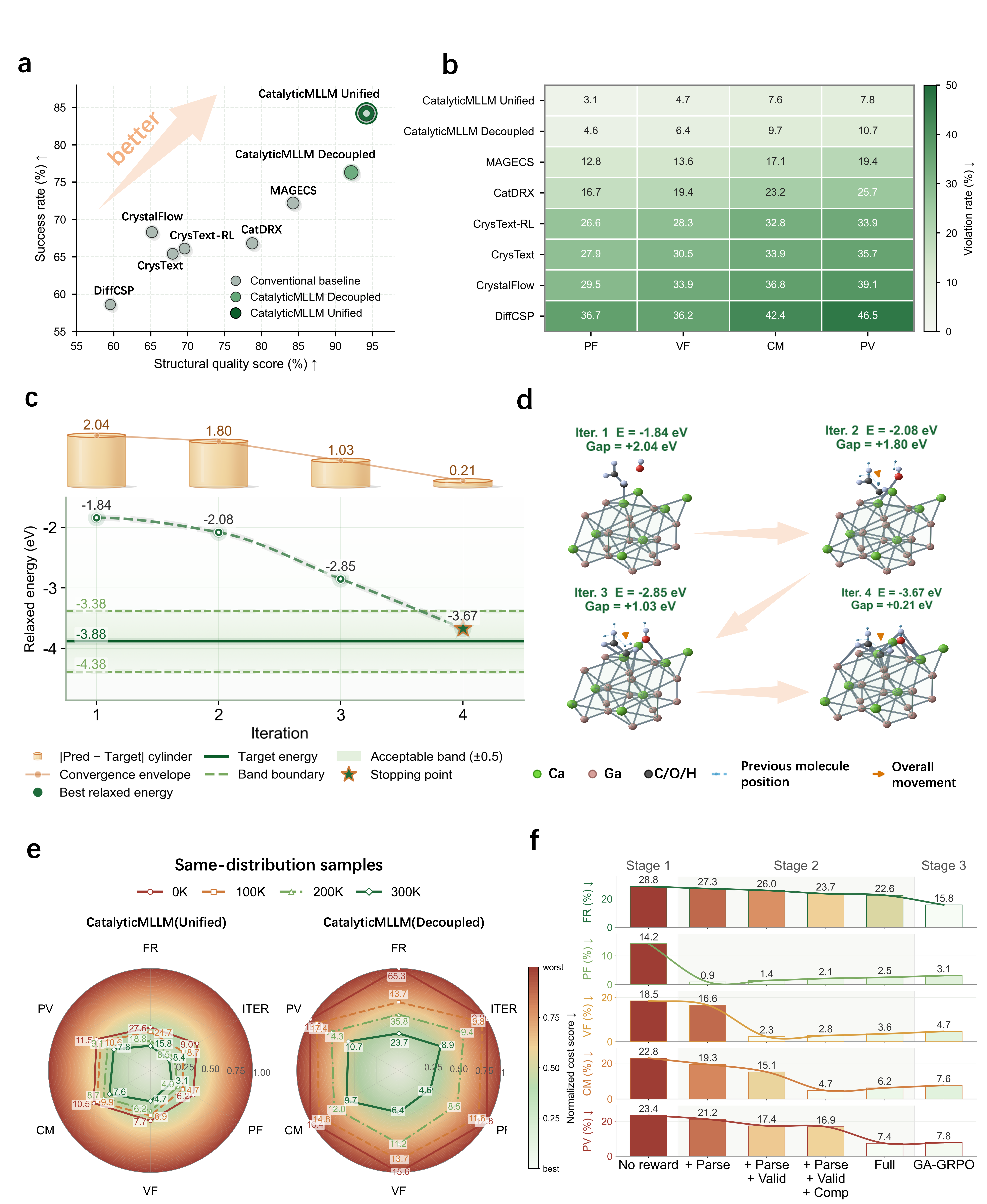}
\caption{
Inverse-design performance evaluation. (a) shows the Pareto plot of the inverse-design success rate and constraint satisfaction rate of different methods. (b) shows the violation rates of four constraints in the CIFs generated by different methods. (c) shows the energy variation of the best CIF generated in each of four iterative optimization rounds during an inverse-design process. (d) shows the structural evolution of the best CIFs generated during the four iterative optimization rounds presented in (c). (e) evaluate the robustness of the unified and separate CatalyticMLLM architectures to data distribution shifts. (f) illustrates the detailed ablation experiments conducted on the reward function. Notes: In panels (b),(e) and (f), PF denotes parse failure; VF, validation failure; CM, composition mismatch; PV, physical violation; FR, failure rate; and ITER, the number of iterations.
}
\label{fig_2}
\vspace{-0.4cm}
\end{figure*}

\subsection{Inverse Design Performance Comparison and Analysis of CatalyticMLLM}
\label{sec:inverse_design_comparison}


To evaluate the performance of CatalyticMLLM on the inverse design task for catalytic materials, we compare it with several representative methods, including DiffCSP, CrystalFlow, CrysText, CrysText-RL, CatDRX, and MAGECS. CatalyticMLLM (Decoupled) is also included as a control to quantify the gains from unified modeling. To ensure a fair comparison, all methods are retrained on the same dataset and evaluated using a consistent protocol.

In the inverse design task, a generated structure is considered successful only if the absolute deviation between its predicted property and the target property is below a predefined threshold. For CatalyticMLLM (Unified), the same model performs both structure generation and property prediction, so relying solely on its own predictions may introduce self-evaluation bias. We therefore additionally validate the generated structures using DFT. A design is counted as successful only if it satisfies the target-property deviation criterion and passes validation by both independent models, ensuring that success is not determined solely by the unified model’s own predictions.

As shown in the Pareto plot in \cref{fig_2}(a), conventional baselines generally exhibit weaker performance in terms of both structural quality and inverse design success rate, whereas the decoupled and unified variants of CatalyticMLLM achieve clear advantages. In particular, CatalyticMLLM (Unified) attains a success rate of $84.2\%$, outperforming CatalyticMLLM (Decoupled), which achieves $76.3\%$. This result indicates that jointly modeling property prediction and inverse design can effectively improve target-property matching.

\cref{fig_2}(b) further compares the structural constraint violation rates of the generated CIF files, including parse failure (PF), validation failure (VF), composition mismatch (CM), and physical violation (PV). Conventional methods generally exhibit high failure rates. For example, all four metrics of DiffCSP exceed $36\%$. Even for the stronger MAGECS baseline, the PF, VF, CM, and PV rates remain $12.8\%$, $13.6\%$, $17.1\%$, and $19.4\%$, respectively. By contrast, the corresponding rates of CatalyticMLLM (Unified) are only $3.1\%$, $4.7\%$, $7.6\%$, and $7.8\%$, demonstrating that its generated structures offer better parseability, compositional consistency, and physical plausibility.

In addition, \cref{fig_2}(c) shows the energy evolution of the predicted CIF during a representative iterative inference process. It can be observed that the target requirement is satisfied after only four rounds of iterative optimization. \cref{fig_2}(d) visualizes the corresponding structural evolution during the iterative process.

In summary, inverse design performance improves progressively from conventional methods to CatalyticMLLM (Decoupled), and further to CatalyticMLLM (Unified). Although conventional methods can generate a certain number of candidate structures, they remain limited in target-property matching and structural-constraint satisfaction. CatalyticMLLM (Decoupled) significantly enhances CIF generation capability by introducing a large language model and reinforcement fine-tuning. CatalyticMLLM (Unified) further integrates property prediction, structure generation, and target-oriented optimization into a single model with a shared representation space, thereby alleviating the representation inconsistency between the predictor and the generator in the decoupled paradigm. These results demonstrate that the unified architecture can generate more stable, parseable, and physically plausible CIF structures, providing a more reliable modeling paradigm for closed-loop inverse design of catalytic materials.

\subsection{Robustness of the Unified and Decoupled Architectures Under Data Distribution Shift}

In practical materials design tasks, property prediction and inverse design data may originate from different distributions, differing, for example, in catalyst composition, adsorbate type, or local structural environment. When the two tasks cover different chemical and structural spaces, the generative model may propose candidate structures outside the reliable applicability domain of the property prediction model. The resulting evaluation errors may affect subsequent screening and propagate through closed-loop optimization. Robustness under data distribution shift is therefore important for ensuring reliable closed-loop inverse design.

The unified architecture jointly learns property prediction and inverse design within a single model, facilitating cross-task knowledge transfer through shared parameters and a shared latent representation space. This mechanism may mitigate representational discrepancies arising from independent training, helping maintain effective coordination between prediction and generation when the task-specific data distributions differ. To evaluate this advantage, we compare the unified and decoupled architectures under varying degrees of data distribution shift.

\subsubsection{Experimental Design}

We construct the experimental data using 300,000 samples, each corresponding to one property prediction example and one inverse design example. We then vary the number of samples in the common-distribution portion of the two tasks' training data to construct different cross-task distribution shift settings. Here, 0K, 100K, 200K, and 300K count the underlying samples, each of which corresponds to a forward--inverse data pair. The settings are defined as follows:

\begin{itemize}
    \item \textbf{0K (large distribution shift)}: The number of samples in the common-distribution portion is approximately zero. Property prediction and inverse design are trained on data drawn from their respective task-specific distributions, representing the largest cross-task distribution discrepancy considered in this experiment.
    
    \item \textbf{100K / 200K (progressively reduced distribution shift)}: The common-distribution portion contains 100k and 200k samples, respectively, while the remaining data are constructed according to the task-specific distributions. As the common-distribution portion increases, the overall distributional discrepancy between the two tasks progressively decreases.
    
    \item \textbf{300K (same-distribution reference)}: The forward and inverse data corresponding to all 300k samples are included in the common-distribution portion, serving as an idealized reference setting without cross-task distribution shift.
\end{itemize}

\subsubsection{Experimental Results and Analysis}

As shown in~\cref{fig_2}(e), the unified architecture consistently outperforms the decoupled architecture overall across the distribution shift settings considered. As the number of common-distribution samples increases from 0K to 300K, cross-task distribution shift progressively decreases, and the overall performance of both architectures improves. However, the decoupled architecture exhibits larger performance variations across settings, whereas the metrics of the unified architecture vary more moderately. In particular, the unified architecture maintains comparatively strong inverse design performance when common-distribution samples are scarce and distribution shift is substantial.

These results indicate that the decoupled architecture is more sensitive to distributional discrepancies between the property prediction and inverse design training data. Because the generative and property prediction models are trained independently, cross-task distributional discrepancies may cause the structural space explored by the generator to diverge from the predictor's reliable applicability domain, thereby affecting closed-loop screening and optimization. By contrast, the unified architecture can exploit structure--property relationships across both tasks through parameter sharing and joint learning. Its more stable performance is consistent with the interpretation that cross-task knowledge transfer helps mitigate the effects of distribution shift.

\subsection{Ablation Analysis of GRPO and GA-GRPO}
\label{subsec:stage23_generalization}

To further verify the effectiveness and generalizability of the last two reinforcement fine-tuning stages proposed in CatalyticMLLM, we conduct an additional set of experiments, with the results presented in \cref{fig_2}(f). Specifically, we further apply the second- and third-stage reinforcement fine-tuning procedures to each original baseline model to evaluate whether these two training stages can consistently improve the inverse design performance across different models.

\subsubsection{Stage 2: Ablation Analysis of GRPO Reward Terms}

The primary objective of Stage 2 is to improve the usability of the generated CIF structures so that they can serve as valid candidates for subsequent property prediction and inverse design. To this end, under identical model initialization and sampling settings, we progressively introduce the Parse, Valid, Comp, and Phys rewards and report the proportions of different failure modes on 2,000 test samples, as shown in \cref{fig_2}(f).

When only SFT is applied, the model exhibits high failure rates across all metrics, with a Parse Fail rate of 14.2\%. This indicates that supervised fine-tuning alone is insufficient to ensure the practical usability of the generated CIF files. After introducing the \textbf{Parse} reward, the Parse Fail rate decreases to 0.9\%, demonstrating that this reward effectively enforces the basic syntactic requirements of the CIF format.

Adding the \textbf{Valid} reward further reduces the Valid Fail rate to 2.3\%, indicating improved field completeness and internal consistency. However, syntactic and structural validity does not necessarily guarantee correct chemical composition. After introducing the \textbf{Comp} reward, the Comp Mismatch rate decreases to 4.7\%, confirming the effectiveness of the compositional consistency constraint.

Finally, under the \textbf{Full} setting, the addition of the physical plausibility reward reduces the Phys Violation rate to 7.4\%, while the other failure rates remain low. This shows that the Phys reward effectively suppresses atomic overlaps and nonphysical configurations.

Overall, the reward terms address complementary failure modes, including syntax errors, incomplete structural fields, compositional mismatches, and geometrically unreasonable configurations. Their progressive introduction improves the generated structures from merely ``parseable'' to ``structurally complete, compositionally consistent, and physically plausible,'' thereby providing a reliable foundation for subsequent closed-loop inverse design.

\subsubsection{Stage 3: Ablation Analysis of GA-GRPO}
\label{subsec:ablation_gpgrro}

After Stage 2 establishes structural usability, Stage 3 further improves the alignment between the generated structures and the target properties. To evaluate the contribution of GA-GRPO, we compare the inverse design success rate and structural constraint metrics before and after introducing this strategy, as shown in \cref{fig_2}(f).

After Stage-2 training, the model can already generate relatively standardized CIF structures with basic physical feasibility. However, the generation process still relies mainly on static distribution learning and lacks the ability to continuously refine structures using target-property feedback. GA-GRPO freezes the property predictor and physical constraint modules, optimizes only the generation policy, and combines group-relative rewards with dynamic exemplar-pool updates to perform a closed-loop ``generation--evaluation--replacement'' search within the local structural neighborhood. This mechanism preserves and exploits high-quality historical candidates, guiding the model progressively toward the target property.

The results show that introducing GA-GRPO further improves the inverse design success rate without significantly degrading structural constraint metrics such as PF, VF, CM, and PV. Thus, the performance gain is not achieved at the expense of structural quality, but instead results from more efficient local search and more stable target alignment.

Therefore, Stage-2 GRPO primarily ensures that the generated structures are ``reasonable and usable,'' whereas Stage-3 GA-GRPO enables further ``target-oriented iterative optimization.'' By continuously refining high-quality candidates, GA-GRPO prevents the model from remaining in feasible but suboptimal structural regions. Together, the two stages enable CatalyticMLLM to achieve high structural reliability, inverse design success rates, and search efficiency.

\section{Conclusion and Discussion}

In this work, we propose CatalyticMLLM, a graph--language multimodal large language model that unifies catalytic-materials property prediction and inverse design. The proposed framework integrates three-dimensional structure encoding, textual semantic modeling, relaxed-energy prediction, and CIF-level structure generation within a single model and a shared representation space. This design mitigates the representation inconsistency, evaluator bias, and error propagation commonly observed in conventional decoupled ``generation--evaluation'' paradigms. Built upon EquiformerV2 and Qwen2.5-VL, CatalyticMLLM can jointly leverage geometric structures and textual information, thereby enabling a closed-loop optimization workflow of ``inverse design--prediction--screening--redesign''.

Experimental results demonstrate that CatalyticMLLM outperforms existing baselines in both property prediction and inverse-design tasks. For property prediction, the model achieves lower MAE and higher $R^2$, indicating that multimodal alignment effectively fuses local geometric information with catalytic semantic knowledge. For inverse design, the unified version consistently surpasses baseline models and the decoupled variant in terms of success rate and related metrics, validating the effectiveness of the unified generation-and-evaluation mechanism.

Further experiments show that inverse-design training can in turn improve property-prediction performance, suggesting that the two tasks are complementary within the shared representation space. The model learns not only the forward mapping from structure to property, but also the inverse constraints from target properties to structural distributions, thereby forming a more stable structure--property representation.

The current method still has certain limitations. At present, CatalyticMLLM is mainly focused on catalytic-materials property prediction and inverse design. In future work, incorporating multi-objective optimization and integrating data from multiple materials domains may further improve the generality and reliability of the model for broader materials discovery.


\section{Method}

\subsection{Architecture Overview}

In this study, we construct \textbf{CatalyticMLLM}, a unified multimodal framework for catalytic materials. The framework integrates \textbf{Equiformer-V2} as the 3D geometric encoder and \textbf{Qwen2.5-VL} as the multimodal large language model (LLM) backbone. Through a trainable linear projection layer, the architecture bridges 3D spatial representations and semantic textual sequences, enabling the model to simultaneously process atomic coordinate information and textual structural descriptors.

\subsection{Three-Stage Training Strategy}

To establish a high-accuracy closed-loop system for property prediction and inverse design, we propose a progressive three-stage training paradigm.

\subsubsection{Stage 1: Multimodal Supervised Fine-Tuning (SFT)}

The goal of this stage is to establish the fundamental mapping relationship between chemical structures and their corresponding energies. We fine-tune the model on a dataset containing approximately 370,000 samlpes, covering the following two core tasks:
\begin{itemize}
    \item \textbf{Property prediction}: mapping $[3D \text{ structure} + \text{text prompt}]$ to the adsorption energy.
    \begin{equation}
        E_{pred} = f_{\theta}(\text{Structure}_{3D}, \text{Prompt}_{text})
    \end{equation}
    \item \textbf{Inverse design}: mapping $[\text{target energy} + \text{text prompt}]$ to a structured CIF file.
\end{itemize}

\subsubsection{Stage 2: Structural Integrity Optimization Based on GRPO}

To address atomic mismatches, such as inconsistent stoichiometry, and nonphysical atomic overlaps in the CIF files generated in the first stage, we introduce GRPO. We design a multi-objective reward function $R_{PVCP}$ to penalize physically unreasonable configurations:
\begin{equation}
    R_{PVCP} = \omega_1 R_{parse} + \omega_2 R_{valid} + \omega_3 R_{comp}+ \omega_4 R_{phys}
\end{equation} 
where $R_{parse}$ denotes CIF parseability, $R_{comp}$ ensures consistency in atomic species and counts, and $R_{valid}$ and $R_{phys}$ represent structural validity and physical plausibility, respectively.
\subsubsection{Stage 3: Iterative reinforcement fine-tuning with GA-GRPO}

In the final stage, we construct an iterative closed-loop inverse design system following a ``design--evaluation--refinement'' workflow. To preserve the physical accuracy of the property predictor, we freeze the Equiformer-V2 encoder, the LLM backbone, and the property prediction head, and optimize only the autoregressive mapping layer using GA-GRPO. Meanwhile, in addition to incorporating property prediction accuracy, this stage strictly inherits the physical constraint terms from Stage 2 to prevent the model from obtaining spurious rewards through nonphysical coordinate manipulation or loopholes in structural representation, thereby ensuring that the optimization process corresponds to genuine structural evolution.

To avoid premature convergence and local optimum locking caused by reliance on a single exemplary structure, we introduce a dynamically updated exemplar pool to perform population-based local search.

Specifically, for each inverse design sample, the model first generates an initial set of candidate CIF files under purely textual conditions,
$\{C_1, C_2, \dots, C_n\}$. The candidate data are evaluated using the frozen multimodal predictor and physical constraint rules, and the top-$K$ structures with the highest scores among those satisfying the hard constraints are selected to form the initial exemplar pool:
 \begin{equation}
\mathcal{P} = \{S_1, S_2, \dots, S_K\}.
\end{equation} 
During the subsequent multimodal refinement process, each iteration randomly data one structure from the exemplar pool $\mathcal{P}$ as the current reference sample. Its 3D geometric features are fed back into the model, together with the target energy and structured textual prompt, to guide the model in generating a new set of candidate CIF files.

A comprehensive reward is computed for the newly generated data:
 \begin{equation}
R = 0.7 R_{energy} + 0.3 R_{PVCP},
\end{equation} 
where the energy-matching reward is defined as:
 \begin{equation}
R_{energy} = \exp(-\lambda |E_{pred}(C_i) - E_{target}|).
\end{equation} 

If any newly generated candidate obtains a score higher than that of the currently lowest-scoring structure in the exemplar pool, it is added to the pool and the worst-performing sample is removed, thereby keeping the pool size constant. The exemplar pool is continuously and dynamically updated throughout the iterative process, enabling population-level structural evolution through a survival-of-the-fittest mechanism.

The above mechanism is equivalent to performing a surrogate-model-guided population-based search within the local structural neighborhood. By randomly sampling exemplars, the method introduces controlled exploration while exploiting high-reward regions and maintaining limited structural diversity, thereby effectively alleviating mode collapse and single-point locking.

This closed-loop process is executed for 10 iterations for each sample, allowing the generative layer, under the physical knowledge constraints fixed by the frozen predictor, to progressively conduct fine-grained search along the local potential energy surface (PES), ultimately achieving stable alignment with the target energy.

\section{Reward Function Design for Reinforcement Fine-Tuning}
\label{sec:grpo_reward}

\paragraph{Stage 2: Reinforcement Fine-Tuning Based on Geometric Plausibility.}
On the basis of the initial generative model obtained from supervised fine-tuning (SFT) in the first stage, we introduce GRPO to perform reinforcement fine-tuning on the generative model obtained from Stage 1. The core objective of this stage is to improve the rationality of the CIF structures generated by the model in terms of three-dimensional geometry and structural quality. Specifically, we parse the generated CIF files to obtain real three-dimensional crystal structures, and introduce several heuristic-rule-based geometric and physical plausibility terms into the reward function to jointly constrain atomic composition consistency, CIF parseability, structural field completeness, and obviously unreasonable geometric configurations. It should be noted that this stage does not impose hard geometric constraints; instead, it guides the model through soft reward terms to gradually favor geometrically more plausible structural distributions while preserving generation diversity. Unlike traditional policy gradient methods that directly rely on absolute rewards, GRPO constructs stable learning signals through group-relative advantages and combines them with KL constraints to limit policy drift, thereby continuously increasing the generation probability of high-quality data while preventing drastic shifts in the generation distribution. This section presents the GRPO objective function, the reward function formulation, and the corresponding implementation details used in this work.

\subsection{Problem Definition and Notation}

Let the conditional input prompt be denoted by $x$, and the CIF text sequence generated by the model be
 \begin{equation}
y = (y_1, y_2, \ldots, y_T),
\end{equation} 
where $y_t$ denotes the $t$-th generated token and $T=|y|$ is the sequence length. The current trainable policy, namely the generative model, is denoted as $\pi_\theta(y|x)$, and the reference policy, namely the frozen SFT model, is denoted as $\pi_{\mathrm{ref}}(y|x)$. During training, we construct a \textbf{group} for each prompt and \textbf{online sample} $K$ candidate outputs from the current policy:
 \begin{equation}
\{y_{1}, y_{2}, \ldots, y_{K}\} \sim \pi_\theta(\cdot|x).
\end{equation} 
For each candidate output $y_k$, its scalar reward $r_k$ is computed by the reward model:
 \begin{equation}
r_k = R(x, y_k) \in \mathbb{R}.
\end{equation} 

\subsection{Reward Function for CIF Generation}

To measure the quality of generated CIF text in terms of engineering usability and structural rationality, we design the reward function as a weighted sum of multiple subterms:
 \begin{equation}
\begin{split}
R(x,y) &= w_{\mathrm{comp}}\,s_{\mathrm{comp}}(x,y)+
w_{\mathrm{parse}}\,s_{\mathrm{parse}}(x,y)+ \\
&\quad w_{\mathrm{valid}}\,s_{\mathrm{valid}}(x,y)+
w_{\mathrm{phys}}\,s_{\mathrm{phys}}(x,y),
\end{split}
\label{eq:reward_linear}
\end{equation} 
where $w_{\mathrm{comp}}+w_{\mathrm{parse}}+w_{\mathrm{valid}}+w_{\mathrm{phys}}=1$. In our experiments, we set
 \begin{equation}
(w_{\mathrm{comp}}, w_{\mathrm{parse}}, w_{\mathrm{valid}}, w_{\mathrm{phys}}) = (0.6, 0.2, 0.1, 0.1).
\end{equation} 

The meanings of the sub-reward terms are as follows.

\paragraph{(1) Atomic Composition Consistency Reward $s_{\mathrm{comp}}$.}
This term measures whether the element types and counts parsed from the generated CIF are consistent with the target composition specified by the prompt. Specifically, the expected element count vector $\mathbf{c}^{\ast}$ is parsed from the prompt, and the actual element count vector $\mathbf{c}$ is parsed from the generated CIF. A composition matching score is then defined accordingly. This term mainly encourages the model to generate structural descriptions that are chemically consistent with the specified catalytic system.

\paragraph{(2) Parseability Reward $s_{\mathrm{parse}}$.}
This term measures whether the generated text can be successfully parsed into a structure object by a standard CIF parser, such as pymatgen or ASE, thereby constraining the output to satisfy basic syntactic formatting requirements. A positive reward is assigned if parsing succeeds; otherwise, the reward is set to $-20$. This reward is primarily used to reduce invalid generations caused by syntax errors.

\paragraph{(3) Structural Validity Reward $s_{\mathrm{valid}}$.}
On the basis of successful parsing, this term further evaluates the completeness and consistency of key fields in the CIF file, including whether lattice parameters, space group information, and atomic-site loop fields are complete, and whether labels are duplicated. This reward encourages the model to generate structural descriptions that conform to the CIF specification, rather than merely producing text that can be parsed.

\paragraph{(4) Physical Plausibility Reward $s_{\mathrm{phys}}$.}
This term is used to suppress obviously unreasonable structures, such as atomic overlaps or extreme unit cells. In implementation, simple geometric heuristic indicators can be computed from the parsed structure, such as the minimum interatomic distance threshold and unit-cell volume range, and then mapped to continuous or piecewise scores.

It should be emphasized that the reward design in Eq.~\eqref{eq:reward_linear} can be adjusted according to the objective of each training stage. For example, in early stages, the weights of parseability and composition consistency can be increased to rapidly improve the proportion of valid data; in later stages, the weights related to physical plausibility can be gradually increased to further improve structural quality.

\subsection{Details of the GRPO Reward Function}
\label{app:grpo_reward}

In this study, Group Relative Policy Optimization (GRPO) is adopted to perform reinforcement fine-tuning of the model. For each input prompt $x$, the model samples a group of candidate outputs online:
 \begin{equation}
    \mathcal{Y}(x) = \{y_1, y_2, \cdots, y_K\},
\end{equation} 
where $K$ denotes the number of responses sampled for each prompt. Each candidate output $y_i$ is a CIF text generated by the model. To evaluate the quality of the generated CIF file, we design a composite reward function tailored to the crystal-structure generation task. The reward function consists of four components: the atomic composition matching reward, the CIF parseability reward, the structural validity reward, and the physical plausibility reward.

The total reward function is defined as:
 \begin{equation}
    R(y) =
    \omega_{\mathrm{comp}} R_{\mathrm{comp}}(y)
    + \omega_{\mathrm{parse}} R_{\mathrm{parse}}(y)
    + \omega_{\mathrm{valid}} R_{\mathrm{valid}}(y)
    + \omega_{\mathrm{phys}} R_{\mathrm{phys}}(y),
\end{equation} 
where the four weights are:
 \begin{equation}
    \omega_{\mathrm{comp}} = 0.60,\quad
    \omega_{\mathrm{parse}} = 0.20,\quad
    \omega_{\mathrm{valid}} = 0.10,\quad
    \omega_{\mathrm{phys}} = 0.10.
\end{equation} 

Therefore, the final reward is:
 \begin{equation}
\begin{aligned}
    R(y) ={}&
    0.60 R_{\mathrm{comp}}(y)
    + 0.20 R_{\mathrm{parse}}(y) \\
    &+ 0.10 R_{\mathrm{valid}}(y)
    + 0.10 R_{\mathrm{phys}}(y).
\end{aligned}
\end{equation} 

The design principles and mathematical definitions of the four reward terms are introduced below.

\subsubsection{Atomic Composition Matching Reward}
\label{app:reward_composition}

The atomic composition matching reward $R_{\mathrm{comp}}$ is the most important component of the overall reward function, with a weight of $60\%$. Its goal is to ensure that both the element types and the number of atoms of each element in the generated CIF file are strictly consistent with the target atomic composition.

Let the target atomic composition be:
 \begin{equation}
    \mathbf{c}^{*} = \{(e, n_e^{*}) \mid e \in \mathcal{E}^{*}\},
\end{equation} 
where $e$ denotes an element symbol, $n_e^{*}$ denotes the expected number of atoms of element $e$ in the target structure, and $\mathcal{E}^{*}$ denotes the set of target elements.

The actual atomic composition parsed from the generated CIF file is:
 \begin{equation}
    \mathbf{c} = \{(e, n_e) \mid e \in \mathcal{E}\},
\end{equation} 
where $n_e$ denotes the actual number of atoms of element $e$ in the generated structure, and $\mathcal{E}$ denotes the set of elements appearing in the generated structure.

We first define the missing-element set and the extra-element set:
 \begin{equation}
    \mathcal{M} = \mathcal{E}^{*} \setminus \mathcal{E},
\end{equation} 
 \begin{equation}
    \mathcal{X} = \mathcal{E} \setminus \mathcal{E}^{*}.
\end{equation} 

Here, $\mathcal{M}$ denotes the elements that should appear in the target structure but are missing from the generated structure, while $\mathcal{X}$ denotes the extra elements that appear in the generated structure but do not exist in the target structure.

If missing elements exist, namely $|\mathcal{M}| > 0$, the generated result is considered to contain a severe error, and the reward is defined as:
 \begin{equation}
    R_{\mathrm{comp}} =
    \max(-15,\; -8 - 2|\mathcal{M}|).
\end{equation} 

If no element is missing but extra elements exist, namely $|\mathcal{M}| = 0$ and $|\mathcal{X}| > 0$, the reward is defined as:
 \begin{equation}
    R_{\mathrm{comp}} =
    \max(-15,\; -5 - 2|\mathcal{X}|).
\end{equation} 

If the element types match exactly, that is,
 \begin{equation}
    \mathcal{E} = \mathcal{E}^{*},
\end{equation} 
we further compare the count error of each element. For each element $e$, the relative error is defined as:
 \begin{equation}
    \epsilon_e =
    \frac{|n_e - n_e^{*}|}{n_e^{*}}.
\end{equation} 

The mean relative error over all elements is:
 \begin{equation}
    \bar{\epsilon}
    =
    \frac{1}{|\mathcal{E}^{*}|}
    \sum_{e \in \mathcal{E}^{*}}
    \frac{|n_e - n_e^{*}|}{n_e^{*}}.
\end{equation} 

Based on the mean relative error, the atomic composition reward is defined as the following piecewise function:
 \begin{equation}
R_{\mathrm{comp}} =
\begin{cases}
10, & \bar{\epsilon} = 0, \\
8, & 0 < \bar{\epsilon} < 0.05, \\
5, & 0.05 \leq \bar{\epsilon} < 0.10, \\
2, & 0.10 \leq \bar{\epsilon} < 0.20, \\
-2, & 0.20 \leq \bar{\epsilon} < 0.30, \\
-5, & 0.30 \leq \bar{\epsilon} < 0.50, \\
-10, & \bar{\epsilon} \geq 0.50.
\end{cases}
\end{equation} 

The core role of this reward term is to constrain the model to generate the correct chemical composition. For CIF generation, even if the structural format is valid, an incorrect element type or atom count means that the structure does not correspond to the target material. Therefore, this term is assigned the highest weight.

\subsubsection{CIF Parseability Reward}
\label{app:reward_parseability}

The CIF parseability reward $R_{\mathrm{parse}}$ is used to determine whether the text generated by the model can be parsed as a valid CIF structure file. This reward term has a weight of $20\%$.

Let $\mathrm{Parse}(y)$ denote the operation of parsing the generated text $y$ as a CIF file. If parsing succeeds, a structure object $s$ can be obtained:
 \begin{equation}
    s = \mathrm{Parse}(y).
\end{equation} 

If parsing succeeds, the parseability reward is:
 \begin{equation}
    R_{\mathrm{parse}}(y) = 10.
\end{equation} 

If parsing fails, the generated text cannot form a valid CIF file, and a strong negative reward is directly assigned:
 \begin{equation}
    R_{\mathrm{parse}}(y) = -20.
\end{equation} 

Thus, the parseability reward can be written as:
 \begin{equation}
R_{\mathrm{parse}}(y) =
\begin{cases}
10, & \text{if } y \text{ can be parsed as a valid CIF}, \\
-20, & \text{otherwise}.
\end{cases}
\end{equation} 

In the implementation, the system first uses professional crystal-structure analysis libraries to parse the CIF text. If parsing fails, the output is considered unsuitable for subsequent structural validation and physical plausibility analysis. Therefore, unparseable CIF outputs are assigned a very low overall reward.

The main function of this reward term is to prevent the model from generating text that only superficially resembles CIF but cannot be read by structural analysis tools. It encourages the model to learn the syntactic structure of CIF files, including lattice parameters, atomic-site information, and required CIF fields.

\subsubsection{Structural Validity Reward}
\label{app:reward_validity}

The structural validity reward $R_{\mathrm{valid}}$ evaluates whether the generated crystal structure is reasonable at the geometric and structural levels. This reward term has a weight of $10\%$.

Unlike the atomic composition reward, the structural validity reward does not directly consider whether the element types and atom counts are correct. Instead, it focuses on whether the generated structure has a reasonable unit cell, coordinates, and density. This reward starts from an initial score of $10$ and subtracts penalties according to the problems identified in the structure:
 \begin{equation}
    R_{\mathrm{valid}} =
    \max(-10,\; 10 - P_{\mathrm{valid}}),
\end{equation} 
where $P_{\mathrm{valid}}$ denotes the structural validity penalty.

The structural validity penalty consists of several components:
 \begin{equation}
    P_{\mathrm{valid}}
    =
    P_{\mathrm{vol}}
    + P_{\mathrm{atom}}
    + P_{\mathrm{coord}}
    + P_{\mathrm{density}}.
\end{equation} 

\paragraph{Unit-cell volume penalty}

Let $V$ denote the unit-cell volume of the generated structure. If the unit-cell volume is too small or too large, the structure is considered abnormal:
 \begin{equation}
P_{\mathrm{vol}} =
\begin{cases}
3, & V < 1.0 \text{ or } V > 100000.0, \\
0, & \text{otherwise}.
\end{cases}
\end{equation} 

This term prevents the model from generating unit cells with extremely abnormal volumes.

\paragraph{Atom-count penalty}

Let $N$ denote the total number of atoms in the generated structure. If the structure contains no atoms, a severe penalty is imposed; if the number of atoms is too large, a moderate penalty is assigned:
 \begin{equation}
P_{\mathrm{atom}} =
\begin{cases}
10, & N < 1, \\
2, & N > 1000, \\
0, & \text{otherwise}.
\end{cases}
\end{equation} 

\paragraph{Fractional-coordinate penalty}

Let the fractional coordinates of the $i$-th atom be:
\begin{equation}
    \mathbf{f}_i = (f_{i1}, f_{i2}, f_{i3}).
\end{equation} 

Under normal conditions, fractional coordinates should lie near $[0,1]$. Considering periodic boundary conditions and numerical errors, this study allows coordinates to fall within the range $[-0.1, 1.1]$. If any coordinate component of an atom lies outside this range, that atom is counted as having abnormal coordinates.

Let $N_{\mathrm{bad}}$ denote the number of atoms with abnormal coordinates:
 \begin{equation}
    N_{\mathrm{bad}}
    =
    \sum_{i=1}^{N}
    \mathbb{I}
    \left[
    \exists j \in \{1,2,3\},
    f_{ij} < -0.1
    \text{ or }
    f_{ij} > 1.1
    \right].
\end{equation} 

The corresponding penalty is:
 \begin{equation}
    P_{\mathrm{coord}}
    =
    \min(3,\; 0.5 N_{\mathrm{bad}}).
\end{equation} 

\paragraph{Density penalty}

Let $\rho$ denote the density of the generated structure. If the density is too low or too high, the structure may be unreasonable:
 \begin{equation}
P_{\mathrm{density}} =
\begin{cases}
2, & \rho < 0.1 \text{ or } \rho > 30.0, \\
0, & \text{otherwise}.
\end{cases}
\end{equation} 

In summary, the structural validity reward can be expressed as:
 \begin{equation}
\begin{aligned}
R_{\mathrm{valid}}
=
\max\big(
-10,\;
10
&- P_{\mathrm{vol}}
- P_{\mathrm{atom}} \\
&- P_{\mathrm{coord}}
- P_{\mathrm{density}}
\big).
\end{aligned}
\end{equation} 

This reward term encourages the model to generate geometrically complete structures with reasonable unit cells, normal coordinate ranges, and non-extreme densities.

\subsubsection{Physical Plausibility Reward}
\label{app:reward_physical}

The physical plausibility reward $R_{\mathrm{phys}}$ is used to check whether the generated structure violates basic physical and chemical principles. This reward term has a weight of $10\%$.

The structural validity reward mainly evaluates whether the CIF structure is formally reasonable, whereas the physical plausibility reward further examines whether the spatial relationships between atoms are physically reasonable. For example, if two atoms are too close to each other, severe atomic overlap may occur; if the average volume per atom is too small or too large, the structure is also likely to be unreasonable.

The physical plausibility reward starts from an initial score of $10$ and subtracts penalties according to the detected issues:
 \begin{equation}
    R_{\mathrm{phys}}
    =
    \max(-10,\; 10 - P_{\mathrm{phys}}),
\end{equation} 
where:
 \begin{equation}
    P_{\mathrm{phys}}
    =
    P_{\mathrm{dist}}
    +
    P_{\mathrm{vpa}}.
\end{equation} 

\paragraph{Minimum interatomic-distance penalty}

Let $d_{ij}$ denote the distance between any two atoms in the generated structure. The minimum interatomic distance is defined as:
 \begin{equation}
    d_{\min}
    =
    \min_{i \neq j} d_{ij}.
\end{equation} 

If $d_{\min}$ is too small, the structure may contain atomic overlap. The penalty term is defined as:
 \begin{equation}
P_{\mathrm{dist}} =
\begin{cases}
8, & d_{\min} < 0.5, \\
3, & 0.5 \leq d_{\min} < 1.0, \\
0, & d_{\min} \geq 1.0.
\end{cases}
\end{equation} 

Here, $d_{\min} < 0.5$ \AA{} indicates severe atomic overlap, while $d_{\min} < 1.0$ \AA{} indicates that atoms are too close to each other.

\paragraph{Volume-per-atom penalty}

Let $V$ denote the unit-cell volume, and let $N$ denote the total number of atoms in the structure. The volume per atom is defined as:
 \begin{equation}
    v_{\mathrm{atom}} = \frac{V}{N}.
\end{equation} 

If the volume per atom is too small, the structure is overly crowded; if it is too large, the structure is overly sparse. The corresponding penalty is:
 \begin{equation}
P_{\mathrm{vpa}} =
\begin{cases}
4, & v_{\mathrm{atom}} < 3.0, \\
2, & v_{\mathrm{atom}} > 150.0, \\
0, & \text{otherwise}.
\end{cases}
\end{equation} 

Therefore, the physical plausibility reward is:
 \begin{equation}
    R_{\mathrm{phys}}
    =
    \max(
    -10,\;
    10
    - P_{\mathrm{dist}}
    - P_{\mathrm{vpa}}
    ).
\end{equation} 

\subsection{Group-Relative Advantage (GRPO)}

Directly using the absolute reward $r_k$ for policy gradient updates may suffer from unstable reward scales and weak comparability across different prompts. GRPO constructs learning signals through \textbf{group-relative advantages}. For the $K$ data under the same prompt, the mean reward within the group is computed as the baseline:
 \begin{equation}
\mu = \frac{1}{K}\sum_{k=1}^{K} r_k.
\end{equation} 
The within-group standard deviation is further computed for normalization to improve numerical stability:
 \begin{equation}
\sigma = \sqrt{\frac{1}{K}\sum_{k=1}^{K}(r_k-\mu)^2}+\epsilon,
\end{equation} 
where $\epsilon$ is a small constant introduced to avoid division by zero. The final advantage is defined as:
 \begin{equation}
A_k = \frac{r_k - \mu}{\sigma}.
\label{eq:advantage}
\end{equation} 
This advantage has a zero-mean property, so the update direction is determined by the \textbf{relative quality} among data within the same group, thereby substantially reducing the impact of reward-scale differences across prompts on training.

\subsection{GRPO Objective Function with KL Constraint}

To prevent the policy distribution from deviating excessively from the Stage-1 SFT model during reinforcement learning, which could lead to language degradation, format drift, or ``exploitative outputs,'' we introduce the reference policy $\pi_{\mathrm{ref}}$ and impose a KL penalty on policy drift. We first define the length-normalized log probability of a sequence under the policy:
 \begin{equation}
\ell_\theta(x,y)=\frac{1}{T}\sum_{t=1}^{T}\log \pi_\theta(y_t|x,y_{<t}),
\label{eq:logprob_theta}
\end{equation} 
and the corresponding reference policy log probability:
 \begin{equation}
\ell_{\mathrm{ref}}(x,y)=\frac{1}{T}\sum_{t=1}^{T}\log \pi_{\mathrm{ref}}(y_t|x,y_{<t}).
\label{eq:logprob_ref}
\end{equation} 

In our implementation, we adopt a \textbf{token-level KL estimate} based on sampled trajectories, averaged over sequence length:
 \begin{equation}
\mathrm{KL}(x,y)=\frac{1}{T}\sum_{t=1}^{T}
\left[
\log \pi_\theta(y_t|x,y_{<t})-\log \pi_{\mathrm{ref}}(y_t|x,y_{<t})
\right].
\label{eq:kl_est}
\end{equation} 
This form is equivalent to a single-trajectory estimate of $\mathrm{KL}(\pi_\theta \Vert \pi_{\mathrm{ref}})$ on the sampled sequence, and in practice it effectively constrains policy drift and improves training stability.

Combining the group-relative advantage with the KL constraint, we define the minimization loss function of GRPO as:
 \begin{equation}
\mathcal{L}(\theta)
=
-\frac{1}{K}\sum_{k=1}^{K}
\left[
A_k \cdot \ell_\theta(x,y_k)
-
\beta \cdot \mathrm{KL}(x,y_k)
\right],
\label{eq:grpo_loss}
\end{equation} 
where $\beta>0$ is the KL penalty coefficient. From an optimization perspective, Eq.~\eqref{eq:grpo_loss} has the following intuitive interpretation:
\begin{itemize}
  \item When a sample $y_k$ obtains a higher reward within the group ($A_k>0$), optimization increases its log probability $\ell_\theta$ under the current policy, thereby increasing the probability of sampling similar high-quality CIFs in the future;
  \item When a sample obtains a lower reward ($A_k<0$), optimization reduces its generation probability and suppresses low-quality outputs;
  \item The KL term penalizes the degree of deviation between the current policy and the reference policy, preventing the model from losing its language capability or formatting constraints during reinforcement learning.
\end{itemize}

In implementation, Eq.~\eqref{eq:grpo_loss} can be decomposed into the following per-sample loss:
 \begin{equation}
\mathrm{loss}(x,y_k)= -A_k\cdot \ell_\theta(x,y_k) + \beta\cdot \mathrm{KL}(x,y_k),
\end{equation} 
which is then averaged over the data within the group.

\subsection{Online Sampling and Training Procedure}

For each parameter update, we perform the following procedure for each prompt in the batch:
\begin{enumerate}
  \item \textbf{Online sampling}: sample $K$ candidate CIFs from the current policy $\pi_\theta(\cdot|x)$ using temperature sampling and top-$p$ truncation.
  \item \textbf{Reward evaluation}: compute the reward $r_k=R(x,y_k)$ for each candidate output, and record the subterm scores for diagnostic purposes.
  \item \textbf{Within-group normalization}: compute the group-relative advantage $A_k$ according to Eq.~\eqref{eq:advantage}.
  \item \textbf{Policy update}: compute Eq.~\eqref{eq:grpo_loss} and perform backpropagation to update $\theta$, while keeping the reference policy $\pi_{\mathrm{ref}}$ frozen.
\end{enumerate}

This online within-group optimization mechanism effectively exploits multiple sampled outputs under the same prompt and forms relative ranking signals, thereby achieving more stable training dynamics and higher sample efficiency in long-text structure generation tasks.

\subsection{Geometric Encoder: EquiformerV2}

Equiformer is a class of graph neural networks satisfying SE(3)/E(3) equivariance, combining equivariant inductive biases with the dynamic modeling capability of Transformers. Its core idea is to replace the scalar operators used in conventional Transformers with equivariant tensor operations on three-dimensional atomic graphs, and to introduce an equivariant graph attention mechanism, thereby flexibly capturing local environmental information while preserving rotation and translation equivariance. EquiformerV2 further improves upon this framework in several aspects: it replaces SO(3) convolution with eSCN convolution, introduces attention re-normalization, separable $S^2$ activation, and separable layer normalization, substantially reducing computational cost under higher-order representations while achieving leading energy and force prediction performance on the S2EF and IS2RE tasks of OC20.

In the training of CatalyticMLLM, we adopt EquiformerV2 pretrained on the OC20 dataset as the geometric encoder for 3D molecular graphs, and extract graph embeddings after the final layer normalization but before the energy/force prediction heads. The model treats each atom as a node. Each node corresponds to a two-dimensional embedding tensor, and the entire system is therefore represented as a three-dimensional tensor. The size of this system-level tensor depends on the number of atoms, the number of spherical harmonic channels, and the maximum degree of the spherical harmonics. Our embedding extraction procedure is as follows: the two-dimensional embedding of each atom is first flattened into a one-dimensional vector, and max pooling is then applied over all atomic vectors to obtain a single system-level embedding. To align it with textual features, we project the embedding through a linear mapping head to the same dimension as the text embedding, which is then used for subsequent geometry--text contrastive learning and multimodal fusion. This design preserves the equivariant geometry-aware capability of EquiformerV2 while enabling geometric features to participate in cross-modal attention computation within the latent space of the LLM in the form of unified vectors.

\subsection{Max--Min Gated Multi-Task Loss (MMTG-Loss)}
\label{subsec:mmtg_loss}

In the first stage of this work, we simultaneously optimize two types of objectives: the regression loss
$L_{\mathrm{MAE}}$ for continuous property prediction, such as the MAE of adsorption energy, and the cross-entropy loss $L_{\mathrm{CE}}$ for discrete-token or generation tasks. A conventional approach typically adopts a linearly weighted form:
 \begin{equation}
    \label{eq:plain_loss}
    \mathcal{L}_{\mathrm{plain}}
    = \lambda\,L_{\mathrm{MAE}} + L_{\mathrm{CE}},
\end{equation} 
where $\lambda>0$ is a manually specified weighting coefficient. However, when the two losses have different numerical scales, or when their convergence rates are not synchronized across different training stages, Eq.~\eqref{eq:plain_loss} often suffers from the problem that
 {one subtask dominates the gradients for an extended period, leading to insufficient learning of the other subtask}.
Moreover, its performance is highly sensitive to the choice of $\lambda$.

To address this issue, we propose a  {Max--Min Tanh-Gated Loss} (MMTG-Loss). The core idea is to let the currently more difficult subtask dominate the optimization, while using the loss of the other subtask as a bounded gating factor to smoothly regulate the overall loss. Specifically, we introduce
 \begin{equation}
    L_{\max} = \max\!\bigl(L_{\mathrm{MAE}},\,L_{\mathrm{CE}}\bigr);
    L_{\min} = \min\!\bigl(L_{\mathrm{MAE}},\,L_{\mathrm{CE}}\bigr)
\end{equation} 
and define the total loss as
 \begin{equation}
    \label{eq:mmtg_loss}
    \mathcal{L}_{\mathrm{MMTG}}
    = L_{\max} + L_{\max}\,\bigl(1 + \lambda\,\tanh(L_{\min})\bigr),
\end{equation} 
where $\lambda\in(0,1]$ is a hyperparameter controlling the gating strength.
Equivalently, Eq.~\eqref{eq:mmtg_loss} can be written as
 \begin{equation}
    \mathcal{L}_{\mathrm{MMTG}}
    = L_{\max}\,\Bigl(2 + \lambda\,\tanh(L_{\min})\Bigr),
\end{equation} 
which shows that $\mathcal{L}_{\mathrm{MMTG}}$ is always proportional to $L_{\max}$ and is modulated by $L_{\min}$ through the bounded function $\tanh(\cdot)$.

Compared with the linearly weighted loss $\mathcal{L}_{\mathrm{plain}}$, MMTG-Loss has the following notable advantages.

\paragraph{(1) Automatically focusing on the currently most difficult subtask.}
By explicitly introducing $L_{\max}=\max(L_{\mathrm{MAE}},L_{\mathrm{CE}})$, the total loss is always dominated by the numerically larger term. Whether in the early or late stages of training, as long as one sub-loss is significantly larger, the corresponding subtask naturally takes over the dominant role in gradient optimization. This avoids the imbalanced learning phenomenon caused by scale mismatch or improper weight selection in linearly weighted formulations.

\paragraph{(2) Smooth and bounded gating modulation through $\tanh(L_{\min})$.}
The subtask corresponding to $L_{\min}$ is relatively ``easier''. We map it to a bounded interval through $\tanh(L_{\min})\in(0,1)$ and construct the gating factor $1-\lambda\tanh(L_{\min})$.
When both losses are large, $\tanh(L_{\min})\approx 1$, and the gating factor approaches a constant, so the model mainly focuses on reducing $L_{\max}$. As training proceeds, $L_{\min}$ gradually decreases and $\tanh(L_{\min})$ becomes smaller. The gating factor then increases accordingly, progressively relaxing the penalty imposed by the overall loss on $L_{\max}$. This yields a training-progress-adaptive optimization strategy that can be interpreted as ``addressing urgent errors first and refining details later.''

\paragraph{(3) Greater robustness to loss scales and reduced hyperparameter sensitivity.}
In a traditional linear combination, if the numerical scales of $L_{\mathrm{MAE}}$ and $L_{\mathrm{CE}}$ differ substantially, $\lambda$ must be carefully tuned; in some cases, it may even need to be re-searched after changing the unit of measurement, for example from eV to meV for energy. In MMTG-Loss, the total loss uses $L_{\max}$ as the reference scale, while $L_{\min}$ only provides relative modulation through the bounded nonlinearity $\tanh(\cdot)$. Therefore, the loss is less sensitive to the absolute scales of the sub-losses and to small perturbations in $\lambda$, exhibiting better stability and transferability in practical training.

\paragraph{(4) Training dynamics better aligned with the needs of multi-task learning.}
From the gradient perspective, $\mathcal{L}_{\mathrm{MMTG}}$ remains monotonically increasing with respect to both sub-losses, while its effective weights adaptively vary across training stages. In the early stage of training, when both losses are large, MMTG-Loss mainly suppresses the larger term. When one subtask has already been learned well, indicated by a significantly reduced loss, the influence of that task is gradually weakened through $L_{\min}$ and $\tanh(L_{\min})$, allowing the optimization process to continuously focus on the objective that has not yet sufficiently converged. This property is conducive to achieving more balanced convergence quality in multi-task scenarios.

In summary, the Max--Min Gated Multi-Task Loss
$\mathcal{L}_{\mathrm{MMTG}}$
introduces a difficulty-aware adaptive weighting mechanism for multi-objective joint optimization while remaining simple to implement. Compared with the conventional linearly weighted loss
$\mathcal{L}_{\mathrm{plain}}$,
it offers clear advantages in stability, robustness, and multi-task balance.

\subsection{Data Format and Structure-to-String Conversion}

\begin{wrapfigure}{r}{0.58\textwidth}
    \centering
    \includegraphics[width=76mm]{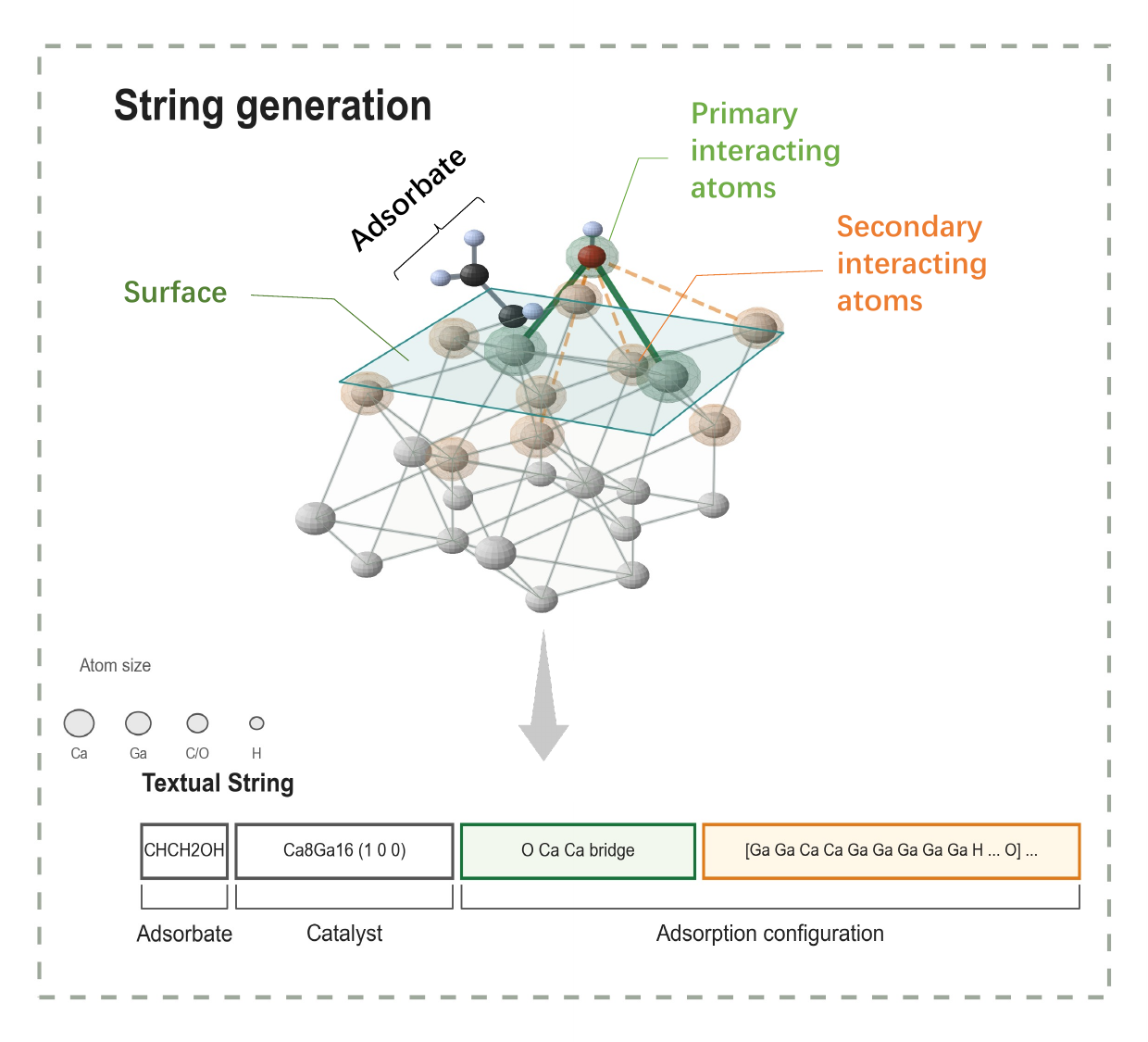}
  \caption{
  Illustration of the three-part textual training data.
  }  
   \label{fig-abt}
    \vspace{-5pt}
  \end{wrapfigure}

The textual inputs in this study strictly follow the three-part string format introduced in the original CatBERTa paper. We convert the structures in the OC20 and OC20-Dense datasets into textual strings, as illustrated in Fig.~\ref{fig-abt}. Each textual input is organized into three segments: the adsorbate, the catalyst surface, and the adsorption configuration.

The  {adsorbate} segment contains only the corresponding elemental symbols. The catalyst surface segment integrates the overall composition of the catalyst and its Miller index, both of which are obtained from the existing metadata of the OC20 dataset.

The third segment, namely the adsorption configuration, is described by identifying the primary and secondary atoms involved in the interaction. This strategy has been validated in previous studies as effective for energy prediction.

In implementation, we use the Pymatgen library to determine the interaction motifs. First, atomic connectivity is constructed according to predefined cutoff radii, where the cutoff radius is determined from the covalent radius of each atom. We then identify atoms connected to the adsorbate atoms and to the top-layer atoms of the surface. Atoms directly connected to the adsorbate atoms are classified as primary interaction atoms, while the neighboring surface atoms of these primary interaction atoms are classified as secondary interaction atoms. Finally, we concatenate the ``adsorbate elemental symbols--catalyst composition and Miller index--primary/secondary nearest-neighbor configuration'' into a structured string, which serves as the standard input format for the textual channel of CatalyticMLLM.

\section{Stage-I Model Training Pipeline}

\subsection{Overall Design}

The training of CatalyticMLLM adopts a three-stage pipeline, following the organizational form of Qwen2.5-VL while incorporating customized designs tailored to the multimodal characteristics of catalytic systems, namely text and 3D atomic graphs. The overall objective is to enable the large language model, while preserving SE(3)/E(3)-equivariant geometric information, to: (i) uniformly represent and exploit the complementary information from text and 3D graphs for energy regression; (ii) maintain robust inference when certain submodalities are missing; and (iii) accomplish bidirectional instruction-following generation and inverse design between energy and structure strings/CIF files.

In terms of notation, we denote the model trained with multimodal data but using only textual inputs during inference as \emph{CatalyticMLLM*}; the model that uses only 3D molecular structures as input, with no catalytic-system information provided in the text, as \emph{CatalyticMLLM$^\Delta$}; and the complete model that uses both geometric and textual inputs during inference as \emph{CatalyticMLLM}.

\subsection{Phase 1: Geometry--Text Alignment Pretraining}

In the initial pretraining stage, CatalyticMLLM mainly trains the molecular 3D structural feature encoder, namely the EquiformerV2 branch, together with the fully connected mapping layer, while keeping the parameters of the LLM frozen. The model uses approximately 100,000 pairs of 3D structures and configuration texts to align the 3D molecular features extracted by EquiformerV2 with the corresponding textual embeddings through cross-modal contrastive learning, thereby establishing geometry--text consistency in the latent space. The goal of this stage is to enable the model to stably capture key geometric information when processing 3D structures and textual data, and to establish one-to-one correspondence with textual descriptions, laying the foundation for subsequent joint pretraining and downstream tasks.

\subsection{Phase 2: Multimodal full-parameter training}

In the second stage, all parameters, including the LLM parameters, are unfrozen on the basis of Stage 1, and training is expanded to a larger-scale pretraining dataset. Compared with Stage 1, this stage introduces approximately 240,000 additional multimodal data. The data cover a broader range, including more adsorbate--catalyst combinations and greater configurational diversity. Compared with the subset used in Stage 1, the dataset in Stage 2 includes and extends the earlier data, thereby improving the model's ability to capture more complex geometric patterns and semantic conditions while maintaining geometry--text alignment.

The training objective of this stage is to fully exploit the complex interactions between 3D molecular structures and textual information within a unified multimodal backbone. In this way, the model can not only achieve accurate relaxed-energy prediction when sufficient geometric information is available, but also fall back to reasonable predictive performance when some modalities are missing. Through joint pretraining on diverse adsorption systems, CatalyticMLLM develops a more comprehensive representational capability for the intricate relationships between 3D structures and text.

\subsection{Phase 3: Multitask Instruction Tuning}

In the instruction-tuning stage, CatalyticMLLM freezes the parameters of EquiformerV2 and the mapping layer and fine-tunes only the LLM component. We construct an instruction dataset containing approximately 360,000 examples, including both multimodal dialogue data and text-only dialogue data. These data cover multiple instruction types, such as energy prediction and inverse generation, in a question-answer format. Through this multidimensional data construction strategy, the model learns how to understand and execute natural-language instructions under different modality combinations, including geometry plus text and text only, thereby exhibiting stronger adaptability and robustness in real-world scenarios with missing modalities and diverse conditions.

\section{Acknowledgments}

\paragraph{Funding:}
This work was supported in part by  Zhongguancun Academy Project No.02012501, and in part by CAS Project for Young Scientists in Basic Research under Grant YSBR-090.

\paragraph{Competing interests:}
All authors of the article have no competing interests.

\bibliography{sn-bibliography}

\newpage
\appendix
\onecolumn
\section*{Appendix For CatalyticMLLM: A Unified Graph--Text Multimodal Large Language Model for Property Prediction and Inverse Design of Catalytic Materials}
\startcontents[app]             
\printcontents[app]{}{1}{\setcounter{tocdepth}{4}} 
\newpage

\section{Algorithmic Procedure of CatalyticMLLM}

This section presents the algorithmic procedures for training and inference in CatalyticMLLM. 
The framework integrates an $E(3)$-equivariant geometric encoder with a multimodal large language model (LLM) to predict adsorption energies, align graph--text representations, and support structure completion through CIF generation.

\subsection{Training Procedure of CatalyticMLLM Step-1}

Algorithm~\ref{alg:qecatalytic_train} summarizes the three-stage training pipeline of CatalyticMLLM First Stage, including structure-to-string preprocessing, graph--text contrastive alignment, multimodal joint optimization with MMTG-Loss, and instruction fine-tuning.\\ \\ \\

\begin{breakablealgorithm}
\caption{CatalyticMLLM First Stage Training Pipeline}
\label{alg:qecatalytic_train}

\small
\setlength{\baselineskip}{11.5pt}

\begin{algorithmic}[1]

\STATE \textbf{Input:} Dataset $\mathcal{D}=\{(G_i,s_i,y_i)\}_{i=1}^{M}$  
\hfill// $G_i$: 3D atomic graph, $s_i$: three-part text, $y_i$: adsorption energy
\STATE Geometric encoder $g_{\phi}$ (EquiformerV2)
\STATE Multimodal LLM $f_{\theta}$
\STATE Projection head $W$, regression head $h_{\mathrm{reg}}$, autoregressive head $h_{\mathrm{ar}}$
\STATE Temperature $\tau$, gating parameter $\lambda$
\STATE \textbf{Output:} Trained parameters $(\phi,\theta,W,h_{\mathrm{reg}},h_{\mathrm{ar}})$

\STATE \vspace{3pt}
\STATE \textbf{Stage 0: Structure-to-String Preprocessing}
\FOR{each valid structure in OC20/OC20-Dense}
    \STATE Convert atomic structure into three-part text
    \hfill// adsorbate / surface composition / adsorption configuration
    \STATE Identify primary and secondary neighbors using Pymatgen
    \hfill// covalent-radius cutoff defines chemical neighborhood
\ENDFOR

\STATE \vspace{3pt}
\STATE \textbf{Stage 1: Graph--Text Contrastive Alignment (Freeze LLM)}
\STATE Freeze $\theta$ to stabilize language representations
\STATE Optimize geometric encoder $g_{\phi}$ and projection $W$
\FOR{each minibatch $\mathcal{B}\subset\mathcal{D}$}
    \FOR{each sample $(G_i,s_i,y_i)\in\mathcal{B}$}
        \STATE $z^g_i \leftarrow g_{\phi}(G_i)$
        \hfill// extract equivariant atomic features
        \STATE $z^g_i \leftarrow \textsc{Flatten}(z^g_i)$
        \STATE $z^g_i \leftarrow \textsc{Pool}(z^g_i)$
        \hfill// max-pool to obtain system-level embedding
        \STATE $\tilde{z}^g_i \leftarrow W z^g_i$
        \hfill// project to text embedding dimension
        \STATE $z^t_i \leftarrow \textsc{TextEnc}(f_{\theta}, s_i)$
        \hfill// encode structured text, no gradient
    \ENDFOR
    \STATE $\mathcal{L}_{\mathrm{align}} \leftarrow
    \textsc{InfoNCE}(\{\tilde{z}^g_i\},\{z^t_i\};\tau)$
    \hfill// enforce cross-modal latent alignment
    \STATE Update $(\phi,W)$ to minimize $\mathcal{L}_{\mathrm{align}}$
\ENDFOR

\STATE \vspace{3pt}
\STATE \textbf{Stage 2: Multimodal Joint Pretraining (MMTG-Loss)}
\STATE Unfreeze all parameters to enable full fusion learning
\FOR{each minibatch $\mathcal{B}\subset\mathcal{D}$}
    \FOR{each $(G_i,s_i,y_i)\in\mathcal{B}$}
        \STATE $\tilde{z}^g_i \leftarrow
        W\cdot \textsc{Pool}(\textsc{Flatten}(g_{\phi}(G_i)))$
        \STATE $H_i \leftarrow \textsc{Fuse}(f_{\theta}; s_i,\tilde{z}^g_i)$
        \hfill// cross-modal Transformer fusion
        \STATE $\hat{y}_i \leftarrow h_{\mathrm{reg}}(H_i)$
        \hfill// adsorption energy regression
        \STATE $L_{\mathrm{MAE}} \leftarrow \lVert \hat{y}_i-y_i\rVert_1$
        \STATE $L_{\mathrm{CE}} \leftarrow -\log p(\text{target tokens}\mid H_i)$
        \hfill// text or structure token prediction
        \STATE $L_{\max} \leftarrow \max(L_{\mathrm{MAE}},L_{\mathrm{CE}})$
        \STATE $L_{\min} \leftarrow \min(L_{\mathrm{MAE}},L_{\mathrm{CE}})$
        \STATE $\mathcal{L}_{\mathrm{MMTG}} \leftarrow
        L_{\max}\cdot(2-\lambda\tanh(L_{\min}))$
        \hfill// adaptive max--min gated multitask loss
    \ENDFOR
    \STATE Update all parameters to minimize $\sum \mathcal{L}_{\mathrm{MMTG}}$
\ENDFOR

\STATE \vspace{3pt}
\STATE \textbf{Stage 3: Instruction Fine-Tuning}
\STATE Freeze geometric encoder and projection layers
\STATE Fine-tune only LLM for instruction-following behavior
\FOR{each instruction minibatch}
    \STATE Format input as natural-language queries
    \hfill// energy prediction or CIF generation prompts
    \STATE Optimize autoregressive cross-entropy loss
\ENDFOR

\STATE \vspace{3pt}
\STATE \textbf{return} trained model parameters

\end{algorithmic}
\end{breakablealgorithm}

\subsection{Training Procedure of CatalyticMLLM Step-2}
Although the Stage-1 supervised fine-tuning endows the model with the basic ability to generate CIF files conditioned on target properties and catalytic-system descriptions, the generated structures may still suffer from syntax errors, incomplete structural fields, composition mismatch, and physically implausible geometries. To improve the structural integrity and practical usability of generated CIF files, we introduce a second-stage reinforcement fine-tuning procedure based on Group Relative Policy Optimization (GRPO).

In this stage, the Stage-1 SFT model is used as the initial policy, while a frozen copy of it is retained as the reference policy. For each inverse-design prompt, the current policy samples a group of candidate CIF files, which are then evaluated by a composite PVCP reward function. This reward function consists of four terms, measuring CIF parseability, structural validity, compositional consistency, and physical plausibility, respectively. The rewards within each sampled group are normalized to obtain group-relative advantages, and the policy is optimized with a clipped objective together with a KL regularization term against the reference policy. In this way, PVCP-GRPO increases the likelihood of structurally reliable CIF generations while preserving the distributional knowledge acquired during supervised fine-tuning.\\

\begin{breakablealgorithm}
\caption{Stage-2 PVCP-GRPO Structural Integrity Optimization}
\label{alg:stage2_pvcp_grpo}

\small
\setlength{\baselineskip}{11.5pt}

\begin{algorithmic}[1]

\STATE \textbf{Input:} Stage-1 SFT policy $\pi_{\theta}$, frozen reference policy $\pi_{\mathrm{ref}}$
\STATE Inverse-design prompt set $\mathcal{D}_{\mathrm{gen}}=\{x_i\}_{i=1}^{N}$
\STATE Number of sampled candidates $K$, KL coefficient $\beta$
\STATE Reward weights $(\omega_{\mathrm{parse}},\omega_{\mathrm{valid}},\omega_{\mathrm{comp}},\omega_{\mathrm{phys}})$
\STATE \textbf{Output:} Stage-2 PVCP-GRPO policy $\pi_{\theta}$

\STATE \vspace{3pt}
\FOR{each minibatch $\mathcal{B}\subset\mathcal{D}_{\mathrm{gen}}$}

    \FOR{each prompt $x_i\in\mathcal{B}$}

        \STATE $\{C_{i,1},\ldots,C_{i,K}\}\leftarrow \pi_{\theta}(\cdot\mid x_i)$
        \hfill// sample a group of CIF candidates

        \FOR{$k=1$ to $K$}
            \STATE $R_{\mathrm{parse}}\leftarrow \textsc{ParseScore}(C_{i,k})$
            \hfill// CIF parseability
            \STATE $R_{\mathrm{valid}}\leftarrow \textsc{ValidScore}(C_{i,k})$
            \hfill// structural field completeness
            \STATE $R_{\mathrm{comp}}\leftarrow \textsc{CompScore}(x_i,C_{i,k})$
            \hfill// composition consistency
            \STATE $R_{\mathrm{phys}}\leftarrow \textsc{PhysScore}(C_{i,k})$
            \hfill// physical plausibility
            \STATE $R_{i,k}\leftarrow \omega_{\mathrm{parse}}R_{\mathrm{parse}}+\omega_{\mathrm{valid}}R_{\mathrm{valid}}+\omega_{\mathrm{comp}}R_{\mathrm{comp}}+\omega_{\mathrm{phys}}R_{\mathrm{phys}}$
            \hfill// PVCP reward
        \ENDFOR

        \STATE $\mu_i\leftarrow \frac{1}{K}\sum_{k=1}^{K}R_{i,k}$,\quad
        $\sigma_i\leftarrow \sqrt{\frac{1}{K}\sum_{k=1}^{K}(R_{i,k}-\mu_i)^2}$
        \hfill// group reward statistics

        \FOR{$k=1$ to $K$}
            \STATE $A_{i,k}\leftarrow (R_{i,k}-\mu_i)/(\sigma_i+\delta)$
            \hfill// group-relative advantage
        \ENDFOR

        \STATE $\mathcal{L}_{i}\leftarrow -\frac{1}{K}\sum_{k=1}^{K}A_{i,k}\log\pi_{\theta}(C_{i,k}\mid x_i)
        +\beta\,\mathrm{KL}\!\left(\pi_{\theta}(\cdot\mid x_i)\,\|\,\pi_{\mathrm{ref}}(\cdot\mid x_i)\right)$
        \hfill// GRPO loss with KL regularization

    \ENDFOR

    \STATE $\mathcal{L}_{\mathrm{Stage2}}\leftarrow \frac{1}{|\mathcal{B}|}\sum_{x_i\in\mathcal{B}}\mathcal{L}_{i}$
    \STATE Update $\theta$ to minimize $\mathcal{L}_{\mathrm{Stage2}}$

\ENDFOR

\STATE \vspace{3pt}
\STATE \textbf{return} Stage-2 PVCP-GRPO policy $\pi_{\theta}$

\end{algorithmic}
\end{breakablealgorithm}

\subsection{Stage-3 GA-GRPO Closed-Loop Inverse Design}

After Stage-2 PVCP-GRPO, the model can generate structurally valid and physically plausible CIF files. However, these candidates may still fail to match the target adsorption energy precisely. Therefore, we further introduce GA-GRPO in Stage 3 to perform target-oriented closed-loop inverse design.

In this stage, CatalyticMLLM first generates an initial set of CIF candidates and evaluates them using its own frozen property-prediction branch together with the PVCP reward. The top-ranked candidates are stored in an exemplar pool. During each refinement step, one exemplar structure is randomly sampled from the pool and used as multimodal guidance for local CIF generation. Newly generated candidates are then evaluated, used to update the exemplar pool through top-$K$ selection, and simultaneously used for GRPO policy optimization. In this way, GA-GRPO combines population-based structural search with reinforcement fine-tuning, enabling the model to progressively approach the target energy while maintaining structural feasibility.

The overall reward used in this stage is defined as:
\[
R(C)
=
0.7R_{\mathrm{energy}}(C)
+
0.3R_{\mathrm{PVCP}}(x,C),
\]
where
\[
R_{\mathrm{energy}}(C)
=
\exp\left(
-\lambda
\left|
f_{\mathrm{pred}}(C,x)-E^{\mathrm{tar}}
\right|
\right).
\]
The complete procedure is summarized in Algorithm~\ref{alg:stage3_ga_grpo}.\\ \\

\begin{breakablealgorithm}
\caption{Stage-3 GA-GRPO Closed-Loop Inverse Design}
\label{alg:stage3_ga_grpo}

\small
\setlength{\baselineskip}{11.5pt}

\begin{algorithmic}[1]

\STATE \textbf{Input:} Stage-2 policy $\pi_{\theta}$, reference policy $\pi_{\mathrm{ref}}$
\STATE Frozen property-prediction branch $f_{\mathrm{pred}}$ in CatalyticMLLM
\STATE Task set $\mathcal{D}_{\mathrm{task}}=\{(x_i,E_i^{\mathrm{tar}})\}_{i=1}^{N}$
\STATE Initial sample size $N_0$, pool size $K$, refinement steps $T$, group size $G$
\STATE KL coefficient $\beta$, energy reward coefficient $\lambda$
\STATE \textbf{Output:} Optimized CIF set $\mathcal{C}^{*}$ and updated policy $\pi_{\theta}$

\STATE \vspace{3pt}
\FOR{each inverse-design task $(x_i,E_i^{\mathrm{tar}})\in\mathcal{D}_{\mathrm{task}}$}

    \STATE $\mathcal{C}_{i}^{0}\leftarrow\{C_{i,1}^{0},\ldots,C_{i,N_0}^{0}\}\sim\pi_{\theta}(\cdot\mid x_i,E_i^{\mathrm{tar}})$
    \hfill// initial text-conditioned CIF generation

    \FOR{each candidate $C\in\mathcal{C}_{i}^{0}$}
        \STATE $\hat{E}(C)\leftarrow f_{\mathrm{pred}}(C,x_i)$
        \hfill// energy prediction by CatalyticMLLM itself
        \STATE $R_{\mathrm{energy}}(C)\leftarrow \exp(-\lambda|\hat{E}(C)-E_i^{\mathrm{tar}}|)$
        \STATE $R(C)\leftarrow 0.7R_{\mathrm{energy}}(C)+0.3R_{\mathrm{PVCP}}(x_i,C)$
        \hfill// target matching and structural feasibility
    \ENDFOR

    \STATE $\mathcal{P}_{i}\leftarrow\textsc{TopK}(\mathcal{C}_{i}^{0},R,K)$
    \hfill// initialize exemplar pool with best feasible CIFs

    \STATE \vspace{3pt}
    \FOR{$t=1$ to $T$}

        \STATE $S_{\mathrm{ref}}\sim\textsc{Uniform}(\mathcal{P}_{i})$
        \hfill// sample exactly one exemplar structure

        \STATE $\mathcal{C}_{i}^{t}\leftarrow\{C_{i,1}^{t},\ldots,C_{i,G}^{t}\}\sim\pi_{\theta}(\cdot\mid x_i,E_i^{\mathrm{tar}},S_{\mathrm{ref}})$
        \hfill// exemplar-guided local generation

        \FOR{$k=1$ to $G$}
            \STATE $\hat{E}_{i,k}^{t}\leftarrow f_{\mathrm{pred}}(C_{i,k}^{t},x_i)$
            \STATE $R_{i,k}^{t}\leftarrow 0.7\exp(-\lambda|\hat{E}_{i,k}^{t}-E_i^{\mathrm{tar}}|)+0.3R_{\mathrm{PVCP}}(x_i,C_{i,k}^{t})$
        \ENDFOR

        \STATE $\mathcal{P}_{i}\leftarrow\textsc{TopK}(\mathcal{P}_{i}\cup\mathcal{C}_{i}^{t},R,K)$
        \hfill// GA-style selection and replacement

        \STATE $\mu_i^t\leftarrow \frac{1}{G}\sum_{k=1}^{G}R_{i,k}^{t}$, \quad
        $\sigma_i^t\leftarrow \sqrt{\frac{1}{G}\sum_{k=1}^{G}(R_{i,k}^{t}-\mu_i^t)^2}$
        \STATE $A_{i,k}^{t}\leftarrow (R_{i,k}^{t}-\mu_i^t)/(\sigma_i^t+\delta)$
        \hfill// group-relative advantage

        \STATE $\mathcal{L}_{i}^{t}\leftarrow -\frac{1}{G}\sum_{k=1}^{G}A_{i,k}^{t}\log\pi_{\theta}(C_{i,k}^{t}\mid x_i,E_i^{\mathrm{tar}},S_{\mathrm{ref}})
        +\beta\,\mathrm{KL}(\pi_{\theta}\|\pi_{\mathrm{ref}})$
        \hfill// reuse candidates for GRPO update

        \STATE Update generative parameters of $\pi_{\theta}$ to minimize $\mathcal{L}_{i}^{t}$

    \ENDFOR

    \STATE $C_i^{*}\leftarrow \arg\max_{C\in\mathcal{P}_{i}}R(C)$
    \hfill// select the best structure from final pool

\ENDFOR

\STATE \vspace{3pt}
\STATE \textbf{return} $\mathcal{C}^{*}=\{C_i^{*}\}_{i=1}^{N}$ and updated policy $\pi_{\theta}$

\end{algorithmic}
\end{breakablealgorithm}

\subsection{Mathematical Properties and Optimization Theory}

Let $a=L_{\mathrm{MAE}}(\theta)$ and $b=L_{\mathrm{CE}}(\theta)$, and define
\[
\textcolor{black}{
\mathcal{L}(a,b)=\max(a,b)\Bigl(2+\lambda\tanh(\min(a,b))\Bigr).
}
\]

\paragraph{Theorem 1 (Monotonicity in Dominant and Secondary Losses).}
For all $a,b \ge 0$:
\begin{enumerate}
    \item $\textcolor{black}{\mathcal{L}(a,b)\text{ is strictly increasing in }L_{\max}.}$
    \item $\textcolor{black}{\mathcal{L}(a,b)\text{ is non-decreasing in }L_{\min}.}$
\end{enumerate}

\paragraph{Proof.}
Assume $a \ge b$. Then
\[
\textcolor{black}{
\mathcal{L}(a,b) = a\Bigl(2 + \lambda \tanh(b)\Bigr).
}
\]
Taking partial derivatives:
\[
\textcolor{black}{
\frac{\partial \mathcal{L}}{\partial a}
= 2 + \lambda \tanh(b) \ge 2,
\quad
\frac{\partial \mathcal{L}}{\partial b}
= a \lambda \operatorname{sech}^2(b) \ge 0.
}
\]
The case $b \ge a$ follows symmetrically. \hfill$\square$

\paragraph{Theorem 2 (Continuity and Piecewise Differentiability).}
$\mathcal{L}(a,b)$ is continuous everywhere on $\mathbb{R}_{\ge 0}^2$, differentiable on regions $\{a>b\}$ and $\{b>a\}$, and subdifferentiable on the boundary $a=b$.

\paragraph{Proof.}
$\max(\cdot)$, $\min(\cdot)$, and $\tanh(\cdot)$ are continuous and Lipschitz. Differentiability holds except at $a=b$, where subgradients exist. \hfill$\square$

\paragraph{Theorem 3 (Local Lipschitz Continuity).}
If $L_{\mathrm{MAE}}(\theta)$ and $L_{\mathrm{CE}}(\theta)$ are locally Lipschitz in $\theta$, then $\mathcal{L}_{\mathrm{MMTG}}(\theta)$ is also locally Lipschitz.

\paragraph{Proof.}
Lipschitz continuity is preserved under composition of Lipschitz functions. \hfill$\square$

\paragraph{Theorem 4 (Gradient Norm Upper Bound).}
Assume $\|\nabla_\theta L_{\mathrm{MAE}}\|\le G_a$ and $\|\nabla_\theta L_{\mathrm{CE}}\|\le G_b$. Then:
\[
\textcolor{black}{
\|\nabla_\theta \mathcal{L}_{\mathrm{MMTG}}\|
\le
\begin{cases}
(2+\lambda)G_a + \lambda L_{\mathrm{MAE}} G_b, & L_{\mathrm{MAE}} \ge L_{\mathrm{CE}},\\[6pt]
(2+\lambda)G_b + \lambda L_{\mathrm{CE}} G_a, & L_{\mathrm{CE}} \ge L_{\mathrm{MAE}}.
\end{cases}
}
\]

\paragraph{Proof.}
Assume $a \ge b$:
\[
\textcolor{black}{
\nabla_\theta \mathcal{L}
= \Bigl(2 + \lambda \tanh(b)\Bigr)\nabla_\theta a
+ a\lambda \operatorname{sech}^2(b)\nabla_\theta b.
}
\]
Since $\textcolor{black}{\tanh(b)\in[0,1)}$ and $\operatorname{sech}^2(b)\le 1$, we obtain
\[
\textcolor{black}{
\|\nabla_\theta \mathcal{L}\|
\le (2+\lambda)G_a + \lambda a G_b.
}
\]
The other case is symmetric. \hfill$\square$

\paragraph{Theorem 5 (Curriculum-Like Optimization Dynamics).}
MMTG-Loss induces an implicit curriculum:
\begin{itemize}
\item \textcolor{black}{When the dominant loss is large, optimization is primarily driven by $L_{\max}$.}
\item \textcolor{black}{The secondary loss $L_{\min}$ acts as a bounded positive gating factor, smoothly modulating the emphasis on $L_{\max}$.}
\item \textcolor{black}{Near convergence, the bounded modulation helps stabilize joint optimization without encouraging the smaller loss to increase.}
\end{itemize}

\paragraph{Proof Sketch.}
\textcolor{black}{Since $\tanh(L_{\min}) \approx 1$ for large $L_{\min}$,}
\[
\textcolor{black}{
\mathcal{L}_{\mathrm{MMTG}} \approx (2+\lambda)L_{\max}.
}
\]
\textcolor{black}{As $L_{\min} \to 0$, the scaling approaches $2L_{\max}$. Thus, the modulation remains bounded and always positive, while preserving the dominant role of $L_{\max}$.} \hfill$\square$

\paragraph{Corollary (Stability vs Linear Weighted Loss).}
Compared with $\mathcal{L}_{\mathrm{plain}} = \lambda L_{\mathrm{MAE}} + L_{\mathrm{CE}}$, MMTG-Loss:
\begin{enumerate}
    \item Is less sensitive to loss scale changes,
    \item Avoids static hyperparameter dominance,
    \item Produces smoother gradient trajectories,
    \item \textcolor{black}{does not introduce a negative coefficient on the smaller loss term,}
    \item Encourages more balanced multitask convergence.
\end{enumerate}

\section{Effect of Different Models on the Results}

To analyze the influence of different model combinations on CatalyticMLLM, we compare text-only models, 3D-structure-only models, and multimodal fusion models. The results are summarized in Table~\ref{tab:model_ablation}. In this table, MAE is used to evaluate the accuracy of adsorption energy prediction, while Success Rate measures the ability of the model to generate candidate structures that satisfy the inverse-design objective. Since property prediction and inverse design correspond to two different mappings, namely ``structure-to-property'' and ``property-to-structure'', we discuss them separately below.

\subsection{Effect on Property Prediction.}
For property prediction, multimodal fusion models consistently outperform single-modality models. The text-only models Qwen2.5 and LLaMA achieve MAEs of $0.565 \pm 0.016$ and $0.597 \pm 0.022$, respectively, indicating that textual descriptions provide useful semantic information but are insufficient to distinguish fine-grained local adsorption configurations. The 3D-structure-only models Equiformer-V2 and SchNet obtain MAEs of $0.591 \pm 0.017$ and $0.892 \pm 0.019$, respectively. Equiformer-V2 clearly outperforms SchNet, suggesting that an equivariant structural encoder has stronger capability in capturing three-dimensional geometric relationships.

After fusing textual and structural information, the prediction error is significantly reduced. Equiformer-V2 + Qwen2.5 achieves the lowest MAE of $0.382 \pm 0.011$, while SchNet + Qwen2.5 reduces the MAE to $0.446 \pm 0.022$, which is much better than using SchNet alone. These results show that textual semantic information and 3D geometric information are complementary for adsorption energy prediction. The 3D encoder provides local geometry and adsorption-configuration information, while the language model contributes system-level descriptions, compositional information, and semantic priors. Their combination enables more effective modeling of the structure--property relationship in catalytic systems.
\begin{wraptable}{r}{0.58\textwidth}
\centering
\caption{Effect of different model combinations on adsorption energy prediction.}
\label{tab:model_ablation}
\resizebox{0.58\textwidth}{!}{
\begin{tabular}{lcc}
\toprule
Model & MAE & Success Rate \\
\midrule
Equiformer-V2 + Qwen2.5 & $0.382 \pm 0.011$ & 84.2 \\
Qwen2.5 & $0.565 \pm 0.016$ & 67.2 \\
Equiformer-V2 & $0.591 \pm 0.017$ & -- \\
SchNet + Llama3.2 & $0.492 \pm 0.018$ & 74.4 \\
Llama3.2 & $0.597 \pm 0.022$ & 62.3 \\
SchNet & $0.892 \pm 0.019$ & -- \\
SchNet + Qwen2.5 & $0.446 \pm 0.022$ & 77.2 \\
Equiformer-V2 + Llama3.2 & $0.487 \pm 0.032$ & 70.6 \\
\bottomrule
\end{tabular}
}
\end{wraptable}
\subsection{Effect on Inverse Design.}
For inverse design, the model combination also has a clear impact on the success rate. When only language models are used, Qwen2.5 and Llama3.2 achieve success rates of $67.2\%$ and $62.3\%$, respectively. This indicates that pure text-based generation can produce a certain proportion of candidate structures, but it lacks explicit 3D geometric constraints and is therefore limited in terms of structural feasibility and target-property matching. For the structure-only models Equiformer-V2 and SchNet, Success Rate is not reported because these models do not possess a complete autoregressive CIF generation capability.

Multimodal fusion models show more stable performance in inverse design. Equiformer-V2 + Qwen2.5 achieves the highest success rate of $84.2\%$, together with the lowest MAE, indicating that this combination can not only predict adsorption energy accurately but also better guide target-conditioned CIF generation. SchNet + Qwen2.5 achieves a success rate of $77.2\%$, outperforming SchNet + LLaMA with $74.4\%$. This suggests that a stronger language model helps improve the stability and target alignment of CIF generation. Under the same structural encoder, Equiformer-V2 + Qwen2.5 also outperforms Equiformer-V2 + LLaMA, further confirming that the capability of the language backbone directly affects inverse-design performance.

\textbf{Overall Analysis.}
Overall, single-modality models show clear limitations in both tasks. Text-only models are insufficient for capturing precise three-dimensional local structural differences, while structure-only models lack language-conditioned modeling and CIF generation capability. Multimodal fusion improves both adsorption energy prediction and target-conditioned structure generation. Among all compared settings, Equiformer-V2 + Qwen2.5 achieves the best results in terms of both MAE and Success Rate, demonstrating that the combination of a strong equivariant 3D structural encoder and a capable language model is most suitable for modeling the structure--semantics--property relationship in catalytic adsorption systems.

\subsection{Analysis of Inverse Design Success Rates under Different Training Data Scales}

\begin{wrapfigure}{r}{0.6\textwidth}
    \centering
    \vspace{-0.2cm}
    \includegraphics[width=0.6\textwidth]{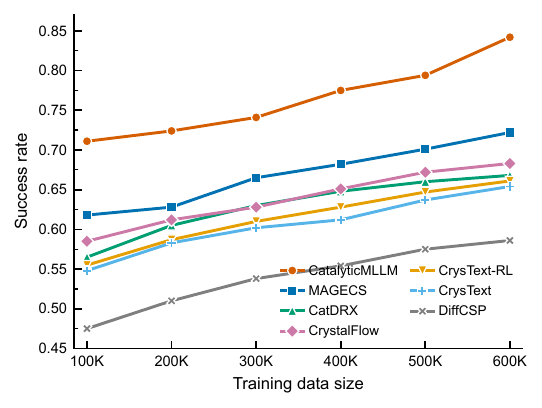}
    \vspace{-0.2cm}
    \caption{Performance trends of the CatalyticMLLM on different sizes of training data}
    \label{fig:data_scale_success_rate}
    \vspace{-0.3cm}
\end{wrapfigure}

To further examine the potential of different algorithms, we evaluated their inverse design performance under varying training data scales. Specifically, we compared DiffCSP, CrysText, CrysText-RL, CrystalFlow, CatDRX, MAGECS, and CatalyticMLLM using 100K, 200K, 300K, 400K, 500K, and 600K training data. The results are shown in Fig.~\ref{fig:data_scale_success_rate}. 

The horizontal axis denotes the training data size, while the vertical axis denotes the inverse design success rate. A higher success rate indicates a stronger capability of generating catalytic material structures that satisfy the specified target-property constraints.

Overall, the inverse design success rates of all methods increase as the training data size grows. This suggests that larger training datasets provide richer structure--property correspondences, thereby improving the ability of models to generate structures under target-property conditions. However, different methods respond differently to the increase in training data size. DiffCSP, CrysText, and CrysText-RL exhibit relatively low success rates under small-data settings. Although their performance improves with more training data, the overall improvement remains limited, indicating that these methods still face challenges in learning the complex mapping between catalytic structures and target properties. In comparison, CrystalFlow, CatDRX, and MAGECS achieve better performance under most data scales, suggesting that stronger generative mechanisms and structural constraints can partially improve inverse design performance.

CatalyticMLLM achieves the highest inverse design success rate across all training data scales, and its advantage is already evident under small-data settings. This demonstrates that CatalyticMLLM has higher data efficiency and can more effectively learn the coupling relationships among compositional information, local structures, and target properties from limited training data. As the training data size further increases, the success rate of CatalyticMLLM continues to improve, indicating that the model not only maintains strong generative capability in small-data scenarios but also consistently benefits from larger-scale datasets, thereby demonstrating favorable scalability.

We attribute this advantage mainly to the unified graph--text multimodal modeling framework of CatalyticMLLM, as well as the strong prior knowledge and representation capacity of large language models.

\section{Energy Prediction under Missing Structural Information and PIR}

\subsection{Indicative CIF Generation}

In practical catalytic materials modeling, incomplete structural information is frequently encountered. For example, for certain adsorption systems, only the adsorbate, catalyst composition, and Miller indices of the exposed surface may be available, whereas the complete three-dimensional atomic structure or local coordination environment is missing. Under such circumstances, accurate relaxed-energy prediction becomes challenging. To improve the applicability of CatalyticMLLM in structure-missing scenarios, we introduce an indicative CIF generation strategy, in which CatalyticMLLM first generates an approximate structural description from limited textual information and then extracts local configurational information from the generated structure for subsequent energy prediction.

Specifically, we provide the model with only the first two segments of the three-part text representation. The first segment contains the chemical symbols of the adsorbate, while the second segment contains the catalyst elemental composition and the Miller indices of the exposed surface. CatalyticMLLM, fine-tuned on the CIF generation task, then autoregressively generates the CIF file. For CatalyticMLLM, this process can be regarded as a conditional text generation task: conditioned on ``adsorbate + catalyst composition + Miller indices'', the model generates a crystal-structure string with a plausible coordination pattern.

It should be noted that, since the input information does not include the true adsorption site or the complete three-dimensional configuration, the CIF files generated by the model are not necessarily physically exact structures. Therefore, we do not directly use the generated CIF as a strict atomic configuration input to the geometric model. Instead, we regard it as an indicative structural representation. Its main role is to provide possible local neighborhood relationships and chemical-environment cues, thereby compensating for the missing configurational segment in the textual input.

Next, we extract atomic coordinates and elemental information from the generated CIF file and convert them into the standard three-part configuration string used by CatalyticMLLM. To improve the robustness of this conversion process against generation errors, we adopt a relatively permissive and simplified neighborhood-extraction strategy. First, we identify the adsorbate atoms closest to the catalyst surface. We then collect the first-shell neighboring atoms around these adsorbate atoms. Furthermore, starting from these first-shell neighbors, we collect the corresponding second-shell neighboring atoms. During neighbor determination, the cutoff radius is set to four times the corresponding covalent radius; that is, atoms within this range are treated as neighboring atoms.

This procedure reflects the core idea of indicative CIF generation. Even if the generated CIF contains deviations in the global crystal structure or precise geometric coordinates, it may still preserve neighborhood information relevant to the local adsorption environment. Since the adsorption energy is strongly influenced by the local coordination relationship between the adsorbate and the surface, the neighborhood information extracted from the generated CIF can still provide useful complementary cues for subsequent energy prediction. Therefore, when the true structural input is unavailable, indicative CIF generation can serve as a structure-completion mechanism, helping CatalyticMLLM recover an approximate local chemical environment from limited textual information.
\begin{figure*}[htp]
    \centering
    \vspace{-5pt}
    \setlength{\belowcaptionskip}{-0.0cm}
       \subfloat[]{
       \includegraphics[width=0.656\linewidth]{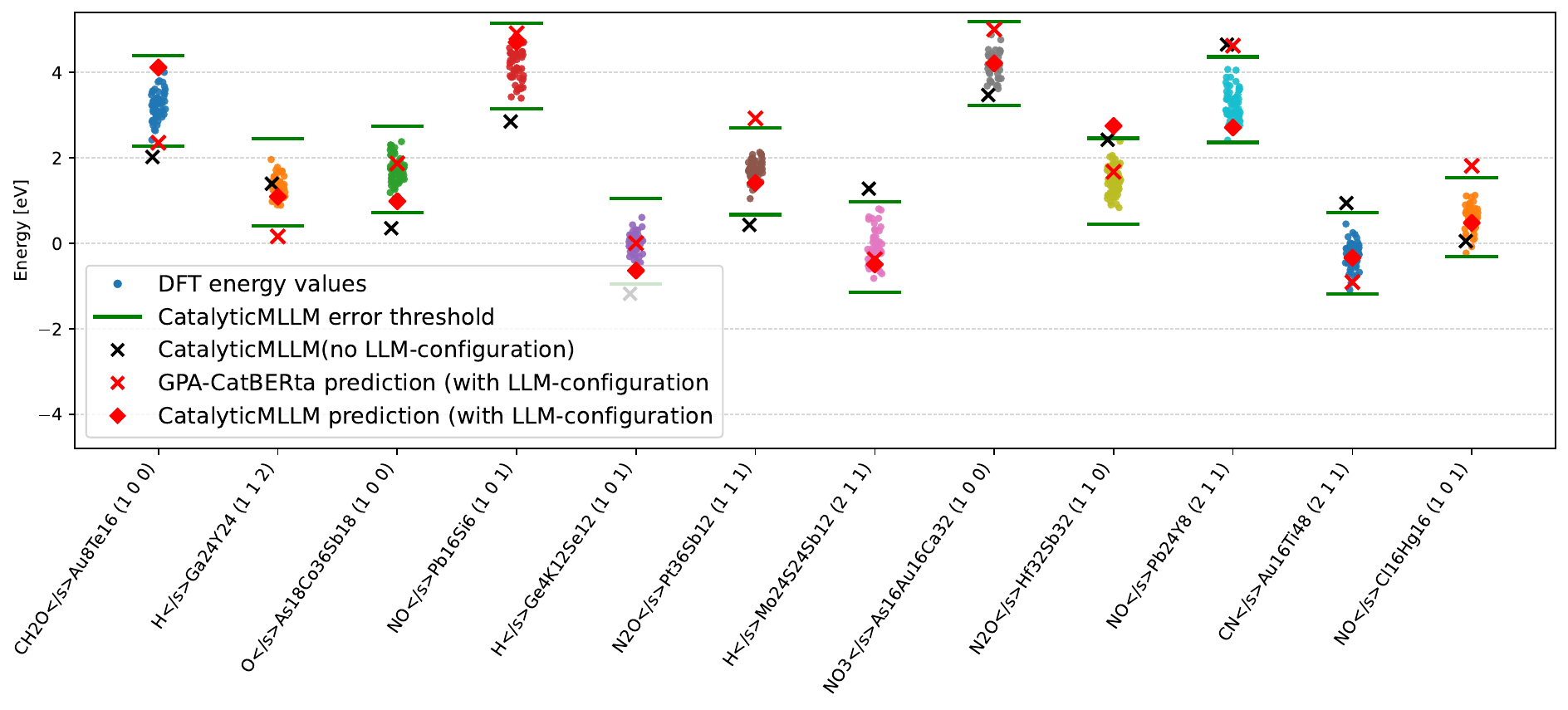} \label{PIRa}}
        \subfloat[]{
        \includegraphics[width=0.335\linewidth]{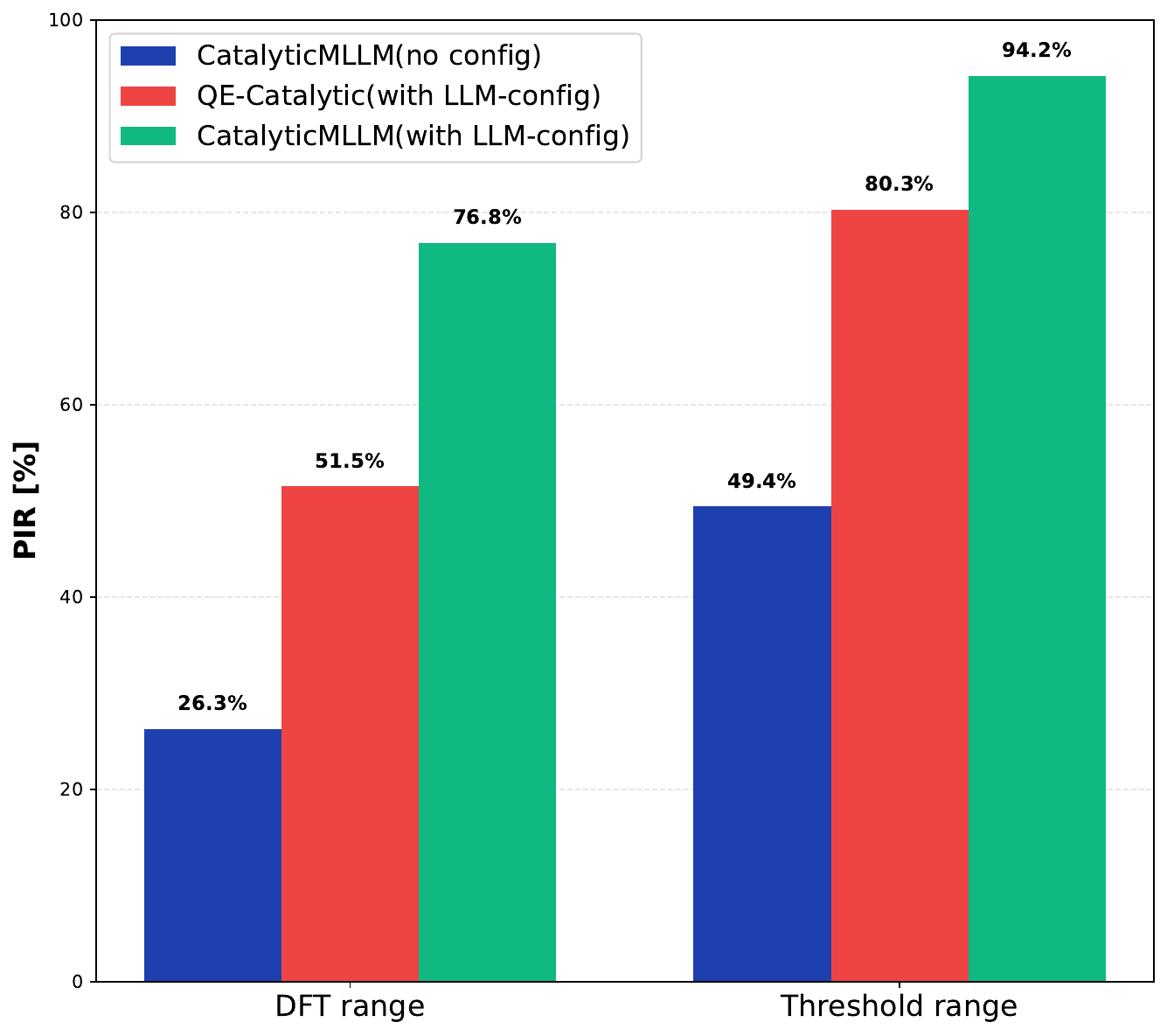}\label{PIRb}}
	  \caption{
   Performance gains from using LLM-derived configuration strings as inputs to CatalyticMLLM. (a) Twelve representative examples selected from 66 adsorbate--catalyst pairs; colored dots indicate energies of different adsorption configurations under each pair. (b) Prediction Inclusion Ratio (PIR) over the 66 pairs, which quantifies the improvement in prediction accuracy after adding LLM-generated configuration strings (``config.'') to the input.}
  
\label{fig3}
\end{figure*}
\subsection{PIR Metric and Experimental Analysis}

To quantitatively evaluate the benefit of indicative CIF generation for energy prediction under missing structural information, we adopt the Prediction Inclusion Ratio (PIR) as the evaluation metric. PIR measures the proportion of model predictions that fall within the target energy range and is defined as follows:
 \begin{equation}
PIR[\%] = \frac{N_{in\text{-}range}}{N_{total}} \times 100,
\end{equation} 
where $N_{in\text{-}range}$ denotes the number of predictions falling within the target energy range, and $N_{total}$ denotes the total number of evaluated data.

In the experiment, we sample 66 adsorbate--catalyst pairs from OC20-Dense and generate multiple candidate CIF files for each pair, resulting in 5,141 possible adsorption configurations. For each candidate configuration, we first extract local neighborhood information from the CIF file generated by CatalyticMLLM and then convert it into a standard three-part configuration string, which is used as the input to CatalyticMLLM for relaxed-energy prediction. This experimental setting simulates a practical application scenario in which structural information is missing but energy evaluation is still required.

The experimental results show that, compared with prediction using only the first two text segments, incorporating the configuration strings generated by CatalyticMLLM significantly improves PIR, with the overall value nearly doubled. This demonstrates that even when the generated CIF files are indicative rather than strictly physically accurate, the local neighborhood chemical information preserved in them can effectively enhance the model's understanding of the adsorption environment, thereby improving the reliability of adsorption energy prediction.

From a methodological perspective, these results indicate that CatalyticMLLM can not only perform multimodal energy prediction with complete structural inputs, but also conduct structural inference and information completion under missing-structure conditions. The model is able to infer possible structural candidates from limited system descriptions and use the local environmental information contained in these candidates to assist property prediction.

\subsection{Multimodal Dataset Construction for adsorption energy Prediction and Inverse Design}
\label{app:dataset_construction}

\paragraph{Motivation and dataset overview.}
To support unified adsorption energy prediction and inverse structural design for catalytic adsorption systems, we construct a large-scale multimodal instruction dataset based on OC20 and OC20-Dense. Existing catalysis datasets are typically organized around single-modality structural representations, such as three-dimensional atomic coordinates, graph representations, or structured textual identifiers. Although these formats are effective for conventional property prediction models, they are not directly suitable for training multimodal large language models that jointly process textual semantics, visual structural information, and three-dimensional atomic geometry.

Therefore, we convert catalytic adsorption structures into instruction-following data with multiple complementary supervision signals. The resulting dataset contains textual descriptions of adsorption systems, CIF-formatted crystal structures, and relaxed-energy labels. This construction allows the model to learn both forward structure--property prediction and inverse property-conditioned structure generation within a unified training framework.

\paragraph{Data sources.}
The dataset is derived from two complementary sources. The first source is a filtered subset of OC20. We start from approximately 460K candidate adsorption systems and retain 345,248 valid structures after quality control. These data mainly provide broad coverage of catalytic surfaces, adsorbates, adsorption sites, and adsorption configurations.

The second source is OC20-Dense, from which we retain 23,999 valid adsorption configurations. Compared with the general OC20 subset, OC20-Dense contains denser local configurational sampling around selected catalytic systems. These data are useful for improving the model's sensitivity to near-degenerate structures and subtle local geometric variations. All OC20-Dense data are checked to avoid overlap with the selected OC20 training subset.

\paragraph{Sample filtering and quality control.}
Before constructing instruction data, all structures are filtered to ensure the reliability of both structural and energetic supervision. We remove data with failed DFT relaxation, non-converged trajectories, corrupted relaxed-energy labels, abnormal adsorption configurations, incomplete atomic coordinates, or physically unreasonable geometries. For relaxation trajectories, only the final converged configuration is retained as the structural input. This filtering procedure ensures that the generated training data are based on physically meaningful structures and valid relaxed-energy annotations.

\paragraph{Task formulation.}
For each valid adsorption system, we construct up to three task-specific instruction data: CIF generation, adsorption energy prediction, and energy-conditioned inverse design. These tasks are designed to jointly cover structural generation, property prediction, and property-driven inverse design.

The three task mappings are summarized as follows:
 \begin{equation}
    \mathrm{Text} \rightarrow \mathrm{CIF},
\end{equation} 
 \begin{equation}
    \mathrm{3D\ Structure} + \mathrm{Text} \rightarrow E_{\mathrm{relaxed}},
\end{equation} 
 \begin{equation}
    E_{\mathrm{target}} + \mathrm{Text} \rightarrow \mathrm{CIF}.
\end{equation} 

The first task teaches the model to generate a structured CIF representation from a textual description of the catalytic system. The second task provides direct supervision for adsorption energy prediction by combining structural and textual inputs. The third task introduces target-energy constraints and trains the model to generate candidate structures conditioned on desired adsorption adsorption adsorption adsorption adsorption adsorption adsorption adsorption adsorption energies. In this way, the dataset explicitly connects the forward mapping from structures to properties with the inverse mapping from properties to structures.

\paragraph{Task-wise sample statistics.}
After task expansion, the final dataset contains 1,078,922 instruction data. The task-wise statistics are reported in Table~\ref{tab:taskwise_counts}. The CIF generation and adsorption energy prediction tasks are constructed for every valid structure, resulting in 369,247 data for each task. The energy-conditioned inverse design task contains 340,428 data. Its sample count is slightly smaller because some systems are removed during target-energy feasibility filtering or do not satisfy the construction criteria for inverse design supervision.

\begin{table}[htbp]
  \centering
  \caption{Task-wise sample statistics of the constructed multimodal dataset.}
  \label{tab:taskwise_counts}
  \begin{tabular}{lrrrr}
    \hline
    Data Source & CIF Generation & adsorption energy Prediction & Inverse Design & Total \\
    \hline
    OC20 Subset & 345{,}248 & 345{,}248 & 318{,}324 & 1{,}008{,}820 \\
    OC20-Dense  & 23{,}999  & 23{,}999  & 22{,}104  & 70{,}102 \\
    \hline
    Total       & 369{,}247 & 369{,}247 & 340{,}428 & 1{,}078{,}922 \\
    \hline
  \end{tabular}
\end{table}

\paragraph{Task 1: CIF generation.}
The CIF generation task, denoted as \texttt{\_cif}, is formulated as a text-to-structure generation problem. The input is a text-only prompt containing the adsorbate, catalyst composition, and Miller index. The output is the corresponding CIF file, including lattice parameters, space group information, atomic species, fractional coordinates, occupancy values, and chemical formula information.

A representative input prompt is:
\begin{quote}
\small
\texttt{CHCH2OH</s>Ca8Ga16 (1 0 0) is the basic identification and chemical formula of a crystal material. Please generate the corresponding CIF (Crystallographic Information File) according to this information.}
\end{quote}

The expected output is a standard CIF-formatted structure:
\begin{quote}
\small
\texttt{Ok, the CIF file I generated is as follows:}\\
\texttt{data\_CHCH2OH</s>Ca8Ga16 (1 0 0)}\\
\texttt{loop\_}\\
\texttt{\_atom\_type\_symbol}\\
\texttt{\_atom\_type\_electronegativity}\\
\texttt{\_atom\_type\_radius}\\
\texttt{\_atom\_type\_ionic\_radius}\\
\texttt{Ca  1.000  1.800  1.140}\\
\texttt{Ga  1.810  1.300  0.760}\\
\texttt{H   2.200  0.250  0.000}\\
\texttt{C   2.550  0.700  0.300}\\
\texttt{O   3.440  0.600  1.260}\\
\texttt{\_symmetry\_space\_group\_name\_H-M   P1}\\
\texttt{\_cell\_length\_a   8.720}\\
\texttt{\_cell\_length\_b   8.901}\\
\texttt{\_cell\_length\_c   30.206}\\
\texttt{\_cell\_angle\_alpha   90.000}\\
\texttt{\_cell\_angle\_beta    90.000}\\
\texttt{\_cell\_angle\_gamma   90.000}\\
\texttt{...}
\end{quote}

This task allows the model to learn CIF syntax, crystallographic formatting, chemical composition consistency, and the structural regularities of catalytic adsorption systems.

\paragraph{Task 2: adsorption energy prediction.}
The adsorption energy prediction task, denoted as \texttt{\_property}, is formulated as a multimodal regression-style instruction task. The input contains textual descriptions of the adsorption system, and optionally three-dimensional structural metadata. The textual description includes the adsorbate, substrate, Miller index, adsorption site, and local coordination environment. The output is the adsorption energy in eV.

A representative input is:
\begin{quote}
\small
\texttt{<image>}\\
\texttt{That's some information about the molecule:}\\
\texttt{CHCH2OH</s>CaGa2 (1 0 0)</s>[O Ca Ca bridge [Ca O] [Ca O]].}\\
\texttt{Please predict its adsorption energy.}
\end{quote}

The corresponding response is:
\begin{quote}
\small
\texttt{The adsorption energy of this molecule is -3.88494148 eV.}
\end{quote}

When available, the three-dimensional structural metadata are also retained:
\begin{quote}
\small
\texttt{"molecule": \{}\\
\texttt{\quad "z": [20, 20, 20, ..., 6, 8],}\\
\texttt{\quad "pos": [[0.36624, 0.0, 14.46867], ...],}\\
\texttt{\quad "cell": [8.72, 8.901, 30.206, 90, 90, 90]}\\
\texttt{\}}
\end{quote}

This task provides direct supervision for quantitative relaxed-energy prediction and encourages the model to align visual, textual, and geometric information.

\paragraph{Task 3: Energy-conditioned inverse design.}
The energy-conditioned inverse design task, denoted as \texttt{\_design\_by\_energy}, is formulated as a target-property-conditioned structure generation problem. The input contains the target adsorption energy and the textual identifier of the catalytic adsorption system. The output is a CIF structure that is expected to satisfy the specified energy constraint.

A representative input prompt is:
\begin{quote}
\small
\texttt{Design a CIF file for the catalytic adsorption system CHCH2OH</s>Ca8Ga16 (1 0 0) with a target adsorption energy of -3.88 eV.}
\end{quote}

The expected output is:
\begin{quote}
\small
\texttt{Ok, the designed CIF file that meets the energy requirement is:}\\
\texttt{data\_CHCH2OH</s>Ca8Ga16 (1 0 0)}\\
\texttt{loop\_}\\
\texttt{\_atom\_type\_symbol}\\
\texttt{\_atom\_type\_electronegativity}\\
\texttt{\_atom\_type\_radius}\\
\texttt{\_atom\_type\_ionic\_radius}\\
\texttt{Ca  1.000  1.800  1.140}\\
\texttt{Ga  1.810  1.300  0.760}\\
\texttt{H   2.200  0.250  0.000}\\
\texttt{C   2.550  0.700  0.300}\\
\texttt{O   3.440  0.600  1.260}\\
\texttt{\_symmetry\_space\_group\_name\_H-M   P1}\\
\texttt{\_cell\_length\_a   8.720}\\
\texttt{\_cell\_length\_b   8.901}\\
\texttt{\_cell\_length\_c   30.206}\\
\texttt{\_cell\_angle\_alpha   90.000}\\
\texttt{\_cell\_angle\_beta    90.000}\\
\texttt{\_cell\_angle\_gamma   90.000}\\
\texttt{...}
\end{quote}

This task provides explicit supervision for inverse materials design. By conditioning generation on target adsorption adsorption adsorption adsorption adsorption adsorption adsorption adsorption adsorption energies, the model learns to associate property requirements with feasible structural configurations.

\paragraph{Discussion.}
The constructed dataset provides a unified data foundation for training multimodal large language models on catalytic materials. Unlike conventional datasets designed for a single prediction objective, our dataset contains both generative and predictive tasks derived from the same adsorption systems. This design enables the model to jointly learn CIF syntax, structural geometry, textual semantics, and relaxed-energy supervision.

More importantly, the dataset supports bidirectional structure--property learning. The adsorption energy prediction task teaches the model to infer properties from structures, whereas the energy-conditioned inverse design task teaches the model to generate structures from target properties. Such bidirectional supervision is essential for closed-loop catalytic materials optimization, where property prediction and inverse design are expected to reinforce each other within a shared representation space.

\section{Baseline Models: Extended Technical Review}

This section presents an in-depth and implementation-level analysis of the baseline models used in adsorption energy prediction and atomic structure modeling. Each model is described with respect to representational design, inductive biases, training methodology, computational cost, physical constraints, and relevance to catalytic datasets such as OC20.

\subsection{CatBERTa}

CatBERTa is a Transformer-based baseline that reframes adsorption configuration learning as a structured language modeling problem. Unlike graph neural networks or equivariant neural architectures that operate on explicit 3D atomic coordinates, CatBERTa encodes adsorption systems as structured textual sequences and processes them using a RoBERTa encoder pretrained on large natural language corpora.

\textbf{Motivation and Conceptual Framework}
The primary motivation behind CatBERTa is to explore whether high-level symbolic and compositional descriptors of catalytic systems contain sufficient predictive signal for adsorption energy estimation. In large-scale datasets such as OC20, adsorption energy depends not only on precise atomic geometry but also on material identity, surface facet type, adsorbate chemistry, and adsorption site category. CatBERTa treats these descriptors as tokens in a domain-specific language, thereby enabling Transformer-based contextual reasoning.

This approach reduces dependence on explicit 3D geometry, enabling learning in scenarios where atomic coordinates are noisy, incomplete, or computationally expensive to process.

\textbf{Input Representation and Tokenization}
Adsorption configurations are serialized into structured text sequences encoding system-level information such as:
\begin{itemize}
    \item Catalyst composition (e.g., transition metal identity)
    \item Crystallographic surface facet (Miller indices)
    \item Adsorbate chemical formula and bonding type
    \item Site classification (top, bridge, hollow, vacancy, defect)
    \item Local coordination descriptors
\end{itemize}

These symbolic descriptors are converted into tokens using subword tokenization, enabling flexible vocabulary expansion for new chemical entities.

\textbf{Model Architecture}
CatBERTa employs a RoBERTa encoder consisting of:
\begin{itemize}
    \item Multi-head self-attention layers for contextual feature aggregation
    \item Positional embeddings to encode sequence ordering
    \item Feed-forward transformation layers for nonlinear feature extraction
    \item Layer normalization and residual connections to stabilize optimization
\end{itemize}

The model outputs a pooled embedding from the [CLS] token, which is passed to a regression head to predict adsorption energy.

\textbf{Training Pipeline}
The training procedure consists of two stages:
\begin{enumerate}
    \item \textbf{Masked Language Modeling (MLM) Pretraining}: Enables the encoder to learn domain-agnostic syntactic and semantic representations.
    \item \textbf{Supervised Energy Fine-tuning}: Uses mean squared error (MSE) loss to regress adsorption energy values.
\end{enumerate}

\textbf{Inductive Bias and Representational Capacity}
CatBERTa imposes minimal physical inductive bias. It does not explicitly enforce rotational or translational invariance and cannot directly model force gradients. As a result, it excels at capturing global compositional trends but struggles with subtle geometric interactions.

\textbf{Computational Properties}
The Transformer backbone enables parallel sequence processing, offering high throughput but quadratic attention complexity in sequence length.

\textbf{Strengths and Limitations}
\textbf{Strengths:}
\begin{itemize}
    \item Operates without explicit atomic coordinates
    \item Efficient inference for large-scale screening
    \item Interpretable token-level attention maps
\end{itemize}

\textbf{Limitations:}
\begin{itemize}
    \item No geometric equivariance
    \item Limited sensitivity to atomic-level distortions
    \item Performance bounded by representational completeness of text encoding
\end{itemize}

\textbf{Relevance to Catalysis}
CatBERTa serves as a baseline for evaluating the predictive capacity of symbolic catalyst representations and is particularly useful in rapid screening pipelines.

\subsection{GAP-CatBERTa}

GAP-CatBERTa (Graph-Augmented Prompt CatBERTa) extends CatBERTa by injecting atomic graph-derived structural priors into the textual prompt, bridging symbolic language models with geometric graph information.

\textbf{Motivation}
Pure text encodings cannot capture subtle variations in atomic neighborhood topology. GAP-CatBERTa addresses this limitation by augmenting the textual input with graph statistics extracted from atomic structure graphs.

\textbf{Graph-Augmented Prompting Mechanism}
The model computes atomic neighborhood features such as:
\begin{itemize}
    \item Elemental neighbor distributions
    \item Coordination numbers
    \item Bond type histograms
    \item Local connectivity metrics
\end{itemize}

These features are serialized into structured text segments appended to the original input prompt.

\textbf{Fusion Strategy}
Graph-derived descriptors are embedded jointly with natural language tokens, allowing Transformer attention to model interactions between symbolic and structural signals.

\textbf{Advantages Over CatBERTa}
\begin{itemize}
    \item Increased sensitivity to local coordination environment
    \item Improved robustness to structural heterogeneity
    \item Better performance on defect-rich catalyst surfaces
\end{itemize}

\textbf{Limitations}
Despite graph augmentation, GAP-CatBERTa remains non-equivariant and lacks explicit force modeling capabilities.

\textbf{Scientific Significance}
GAP-CatBERTa represents a transitional paradigm between pure symbolic modeling and physics-aware graph learning.

\subsection{GemNet-OC}

GemNet-OC is a state-of-the-art directional message-passing graph neural network optimized for the Open Catalyst (OC20) dataset. It explicitly models angular interactions and three-body atomic correlations to achieve high-fidelity adsorption energy prediction.

\textbf{Design Philosophy}
GemNet-OC extends conventional message passing neural networks by incorporating directional geometric constraints that capture bond-angle dependencies crucial for surface chemistry.

\textbf{Graph Construction}
Each atomic system is represented as a graph:
\begin{itemize}
    \item Nodes correspond to atoms
    \item Edges connect atoms within a cutoff radius
    \item Edge attributes encode interatomic distances
\end{itemize}

\textbf{Directional Message Passing}
GemNet propagates messages along triplet interactions $(i,j,k)$, enabling the network to encode angular relationships via:
\begin{itemize}
    \item Distance-based radial basis expansions
    \item Angle-dependent feature modulation
\end{itemize}

\textbf{Mathematical Formulation}
Message updates are parameterized as:
\[
m_{ij}^{(l+1)} = \sum_{k \in \mathcal{N}(j)} \phi(d_{ij}, d_{jk}, \theta_{ijk})
\]
where $\theta_{ijk}$ denotes the bond angle.

\textbf{Physical Inductive Bias}
GemNet-OC enforces:
\begin{itemize}
    \item Rotational and translational invariance
    \item Smooth distance-based interaction kernels
    \item Multi-body interaction expressiveness
\end{itemize}

\textbf{Training Objective}
The model is trained using MSE loss for adsorption energy and optionally force supervision.

\textbf{Performance Characteristics}
GemNet-OC achieves strong accuracy on OC20 while maintaining reasonable computational efficiency compared to fully equivariant Transformers.

\textbf{Strengths}
\begin{itemize}
    \item Excellent accuracy on catalytic adsorption tasks
    \item Captures subtle angular interactions
    \item Strong physical interpretability
\end{itemize}

\textbf{Limitations}
\begin{itemize}
    \item Higher computational cost than SchNet
    \item Requires carefully tuned cutoff radii and basis resolution
\end{itemize}

\textbf{Role in Benchmarking}
GemNet-OC is widely considered a gold-standard GNN baseline for adsorption energy prediction.

\subsection{SchNet}

SchNet is a continuous-filter convolutional neural network designed for atomistic simulations and molecular property prediction. It was among the first deep learning models to introduce continuous spatial filters for learning atomic interactions in 3D space.

\textbf{Core Motivation}
Classical graph neural networks discretize interactions into edge categories, which limits their ability to represent smooth physical potentials. SchNet addresses this limitation by parameterizing interatomic interactions using continuous radial basis filters that operate on Euclidean distances.

\textbf{Atomic Representation}
Each atom is initialized with a learnable embedding vector determined by its element type. Atomic neighborhoods are constructed using a distance cutoff, ensuring locality while enabling scalability to large systems.

\textbf{Continuous-Filter Convolutions}
Interaction blocks compute messages as:
\[
m_{ij} = h_j \cdot W(d_{ij})
\]
where $W(d)$ is a learned continuous filter expressed using radial basis functions. This enables smooth distance-dependent interactions that approximate physical force fields.

\textbf{Update Mechanism}
Messages are aggregated and used to update atomic embeddings through residual transformations, allowing deep stacking of interaction layers.

\textbf{Inductive Bias}
SchNet enforces:
\begin{itemize}
    \item Rotational invariance
    \item Translational invariance
    \item Continuous distance smoothness
\end{itemize}

However, it does not explicitly model angular dependencies, limiting expressiveness for highly directional chemical bonding.

\textbf{Training and Outputs}
The model is typically trained with MSE loss on total energy, with optional force supervision derived from energy gradients.

\textbf{Strengths and Limitations}
\textbf{Strengths:}
\begin{itemize}
    \item Computational efficiency
    \item Stable training dynamics
    \item Smooth and physically meaningful interaction kernels
\end{itemize}

\textbf{Limitations:}
\begin{itemize}
    \item Lacks explicit angular modeling
    \item Lower accuracy on anisotropic surface interactions
\end{itemize}

\textbf{Relevance to Catalysis}
SchNet serves as a foundational baseline, providing a strong reference point for continuous distance-based modeling in surface chemistry.

\subsection{PaiNN}

PaiNN (Polarizable Atomic Interaction Neural Network) is an E(3)-equivariant neural network that propagates both scalar and vector-valued atomic features to model directional polarization effects in molecular systems.

\textbf{Key Insight}
Many chemical interactions depend on directional polarization and anisotropic electron distributions. PaiNN captures these effects by introducing vector channels that transform equivariantly under 3D rotations.

\textbf{Feature Structure}
Each atom maintains:
\begin{itemize}
    \item Scalar embeddings representing isotropic chemical identity
    \item Vector embeddings encoding directional field information
\end{itemize}

\textbf{Equivariant Message Passing}
Messages are constructed using distance-based filters and directional unit vectors:
\[
\vec{m}_{ij} = f(d_{ij}) \cdot \hat{r}_{ij}
\]

This ensures that vector features rotate consistently with atomic geometry.

\textbf{Update Dynamics}
Scalar and vector channels are updated through coupled transformations, enabling modeling of induced dipoles and polarization-like effects.

\textbf{Physical Consistency}
PaiNN enforces strict E(3)-equivariance, making it well-suited for force prediction and molecular dynamics simulations.

\textbf{Advantages}
\begin{itemize}
    \item Strong directional modeling capability
    \item Accurate force field prediction
    \item Better generalization to rotated structures
\end{itemize}

\textbf{Limitations}
\begin{itemize}
    \item Higher computational cost than SchNet
    \item Less expressive than higher-order tensor models
\end{itemize}

\textbf{Impact on Catalysis Benchmarks}
PaiNN provides a competitive middle ground between efficiency and geometric expressiveness in adsorption energy prediction.

\subsection{DimeNet++}

DimeNet++ is a directional message-passing neural network that explicitly models bond-angle interactions using spherical Bessel functions and spherical harmonics.

\textbf{Scientific Motivation}
Two-body distance interactions alone cannot fully capture the angular dependence of chemical bonding. DimeNet++ introduces explicit three-body interaction modeling to address this limitation.

\textbf{Radial and Angular Basis Expansion}
Interatomic distances are expanded using radial basis functions, while bond angles are encoded using spherical harmonics:
\[
\phi_{ijk} = R(d_{ij}) \cdot Y_l(\theta_{ijk})
\]

\textbf{Triplet Message Passing}
Messages are passed over atom triplets, enabling modeling of angular correlations critical for catalytic surface reactions.

\textbf{Efficiency Improvements Over DimeNet}
DimeNet++ reduces computational overhead via:
\begin{itemize}
    \item Reduced basis size
    \item Shared interaction blocks
    \item Optimized memory usage
\end{itemize}

\textbf{Accuracy Characteristics}
DimeNet++ achieves high accuracy on molecular energy benchmarks, particularly where angular bonding is critical.

\textbf{Trade-offs}
\begin{itemize}
    \item Excellent geometric expressiveness
    \item Higher computational cost than GemNet
\end{itemize}

\textbf{Benchmark Role}
DimeNet++ remains a canonical baseline for evaluating three-body geometric modeling.

\subsection{Equiformer}

Equiformer is an E(3)-equivariant Transformer architecture that extends the self-attention mechanism to operate on 3D atomic systems while rigorously preserving Euclidean symmetry. It integrates group representation theory into Transformer design, enabling physically consistent modeling of atomic interactions under arbitrary rotations and translations.

\textbf{Motivation and Conceptual Contribution}
Conventional Transformers operate on permutation-invariant sequences and lack inductive bias for geometric symmetry. However, atomic systems obey E(3) symmetry, requiring model outputs to transform consistently under rigid body transformations. Equiformer addresses this by embedding atomic features into irreducible representations (irreps) of the SO(3) rotation group and replacing scalar attention with equivariant tensor-product attention.

This design enables Equiformer to unify the expressive power of Transformers with the physical constraints of atomistic modeling, improving generalization across molecular conformations and surface orientations.

\textbf{Equivariant Representation Space}
Each atomic embedding is decomposed into multiple irreps labeled by angular momentum order $l$:
\[
\mathbf{h}_i = \bigoplus_{l=0}^{L} \mathbf{h}_i^{(l)}
\]
where:
\begin{itemize}
    \item $l=0$ channels represent scalar features (e.g., element identity)
    \item $l=1$ channels encode vectorial geometric information
    \item $l>1$ channels capture higher-order tensor features
\end{itemize}

This hierarchical representation allows multi-resolution encoding of local and global geometric structure.

\textbf{Equivariant Self-Attention Mechanism}
Unlike standard dot-product attention, Equiformer computes attention through tensor product coupling between irreps:
\[
\mathbf{A}_{ij}^{(l)} = \sum_{l_1, l_2} C_{l_1 l_2}^{l} \left( \mathbf{q}_i^{(l_1)} \otimes \mathbf{k}_j^{(l_2)} \right)
\]
where $C_{l_1 l_2}^{l}$ denotes Clebsch--Gordan coefficients enforcing angular momentum coupling rules.

This ensures that attention outputs transform equivariantly under rotations:
\[
\mathbf{h}_i' = R \cdot \mathbf{h}_i
\]

\textbf{Message Passing and Geometric Encoding}
Equiformer incorporates relative positional encoding based on spherical harmonics $Y_l(\hat{r}_{ij})$ and radial basis expansions:
\[
\phi_{ij}^{(l)} = f(d_{ij}) \cdot Y_l(\hat{r}_{ij})
\]
enabling explicit modeling of angular and distance-dependent atomic interactions.

\textbf{Physical Inductive Bias}
Equiformer enforces:
\begin{itemize}
    \item Exact rotational equivariance
    \item Translational invariance
    \item Distance-based locality
    \item Smooth energy manifolds
\end{itemize}

This ensures physically meaningful force predictions and robust generalization across rotated catalyst surfaces.

\textbf{Computational Complexity}
The dominant computational cost arises from tensor contractions:
\[
\mathcal{O}(N^2 L^3)
\]
where $N$ is the number of atoms and $L$ is the maximum angular momentum order. While global attention improves long-range reasoning, it imposes significant memory and FLOP overhead compared to message-passing GNNs.

\textbf{Strengths}
\begin{itemize}
    \item Captures long-range many-body atomic interactions
    \item Strong generalization across rotated and distorted structures
    \item High accuracy on OC20 adsorption energy and force benchmarks
\end{itemize}

\textbf{Limitations}
\begin{itemize}
    \item High computational and memory cost
    \item Complex implementation due to SO(3) tensor algebra
    \item Training instability for deep high-order tensor stacks
\end{itemize}

\textbf{Scientific and Benchmark Significance}
Equiformer represents a milestone in symmetry-aware deep learning by demonstrating that Transformer architectures can be extended into fully equivariant physical modeling frameworks, setting a new reference for high-capacity atomistic neural networks.

\subsection{EquiformerV2}

EquiformerV2 is an optimized second-generation E(3)-equivariant Transformer that significantly improves computational efficiency, numerical stability, and scalability while preserving the symmetry-aware attention mechanisms of Equiformer.

\textbf{Motivation for Redesign}
While Equiformer achieves strong accuracy, its cubic tensor-product complexity limits scalability to large atomic systems such as OC20 slabs. EquiformerV2 introduces architectural refinements that reduce tensor ranks, factorize attention kernels, and optimize memory layout, enabling training on larger structures and datasets.

\textbf{Factorized Equivariant Attention}
EquiformerV2 decomposes high-order tensor interactions into low-rank factorized components:
\[
\mathbf{A}_{ij}^{(l)} \approx \sum_{r=1}^{R} \mathbf{u}_r^{(l)}(i) \mathbf{v}_r^{(l)}(j)
\]
This reduces computational complexity from cubic to approximately quadratic in angular momentum order.

\textbf{Reduced Irrep Bandwidth}
The model strategically limits the maximum irrep order $L$, prioritizing lower-order channels that contribute most to energy prediction accuracy. This significantly reduces compute without sacrificing representational fidelity.

\textbf{Improved Numerical Stability}
EquiformerV2 incorporates:
\begin{itemize}
    \item Norm-preserving tensor normalization
    \item Residual scaling to stabilize deep equivariant stacks
    \item Improved initialization schemes for Clebsch--Gordan tensors
\end{itemize}

These improvements mitigate gradient explosion and improve training convergence on large datasets.

\textbf{Memory and Throughput Optimization}
Engineering optimizations include:
\begin{itemize}
    \item Kernel fusion for tensor operations
    \item Reduced precision computation (FP16/BF16)
    \item Sparse neighborhood masking to reduce quadratic attention cost
\end{itemize}

These changes enable multi-million structure training on OC20-scale datasets.

\textbf{Performance Gains}
EquiformerV2 achieves state-of-the-art performance on adsorption energy, force prediction, and structure relaxation tasks, while reducing training time and GPU memory consumption by up to 2--4$\times$ relative to Equiformer.

\textbf{Comparative Advantages}
\begin{itemize}
    \item Comparable accuracy to Equiformer at lower computational cost
    \item Improved scalability to large catalyst slabs and defect-rich surfaces
    \item Better robustness to noisy atomic coordinates
\end{itemize}

\textbf{Remaining Challenges}
\begin{itemize}
    \item Still computationally heavier than message-passing GNNs
    \item Complex equivariant kernel engineering
    \item Limited interpretability of high-order tensor channels
\end{itemize}

\textbf{Role in Catalysis Benchmarks}
EquiformerV2 is currently regarded as one of the strongest high-capacity baselines for OC20 adsorption energy prediction, bridging the gap between geometric accuracy and scalable Transformer architectures.

\subsection{UMA}

UMA, short for \emph{Universal Models for Atoms}, is a family of large-scale atomistic foundation models designed for general-purpose prediction across molecules, materials, and catalytic systems. Unlike task-specific neural network potentials trained on a single dataset or chemical domain, UMA is pretrained on large-scale three-dimensional atomic structures collected from multiple domains, enabling broad transferability and strong zero-shot or few-shot generalization.

\textbf{Motivation}
Accurate atomistic modeling usually requires specialized machine-learning potentials trained for a specific dataset, level of theory, or material class. However, such models often generalize poorly when transferred to new chemical environments. UMA addresses this limitation by scaling both model capacity and training data diversity, aiming to construct a universal atomistic model that can provide reliable energy- and force-related predictions across heterogeneous systems.

\textbf{Universal Atomistic Representation}
UMA represents an atomic system as a geometric graph:
\[
\mathcal{G} = (\mathcal{V}, \mathcal{E}),
\]
where atoms are treated as nodes and local interatomic neighborhoods are represented as edges. Atomic numbers, positions, and pairwise geometric information are encoded into symmetry-aware representations. The model preserves fundamental physical symmetries, including translation, rotation, and permutation equivariance or invariance, which are essential for consistent energy and force prediction.

\textbf{Mixture of Linear Experts}
A key architectural component of UMA is the \emph{Mixture of Linear Experts} (MoLE) design. Instead of activating all model parameters for every atomic structure, UMA routes representations through a subset of lightweight linear experts:
\[
\mathbf{h}'_i = \sum_{k=1}^{K} \alpha_{ik} \mathbf{W}_k \mathbf{h}_i,
\]
where $\mathbf{h}_i$ denotes the atomic representation, $\mathbf{W}_k$ is the $k$-th linear expert, and $\alpha_{ik}$ is the routing weight. This design increases model capacity while keeping the number of active parameters and inference cost manageable.

\textbf{Scalability and Generalization}
By training on a large and chemically diverse corpus of atomic structures, UMA learns transferable representations that cover broad chemical environments, including molecular systems, crystalline materials, and catalytic surfaces. This makes it a strong baseline for evaluating whether task-specific or multimodal models can outperform large pretrained atomistic foundation models on adsorption-energy prediction.

\textbf{Comparative Advantages}
\begin{itemize}
    \item Broad transferability across molecules, materials, and catalysts
    \item Strong scalability through large-scale pretraining
    \item Efficient capacity scaling via mixture-of-linear-experts modules
    \item Physically consistent modeling of atomic structures under Euclidean symmetries
\end{itemize}

\textbf{Remaining Challenges}
\begin{itemize}
    \item High pretraining cost due to the large model and dataset scale
    \item Limited task-specific interpretability for catalytic adsorption mechanisms
    \item Dependence on the coverage and quality of the pretrained structural corpus
    \item Weaker integration of symbolic or textual catalytic knowledge compared with multimodal graph--language models
\end{itemize}

\textbf{Role in Catalysis Benchmarks}
In this work, UMA is used as a strong universal atomistic baseline. Its performance reflects the capability of large pretrained geometric models for adsorption-energy prediction, while comparison with CatalyticMLLM highlights the additional benefits of incorporating structured textual information and graph--language alignment.

\subsection{E2GNN}

$E^{2}$GNN is an efficient equivariant graph neural network designed for interatomic potential and force prediction in molecular and crystalline systems. Its main objective is to preserve geometric equivariance while reducing the computational cost typically associated with high-order tensor representations, spherical harmonics, or expensive tensor-product operations.

\textbf{Motivation}
Many equivariant neural networks improve prediction accuracy by explicitly modeling high-order geometric features. However, these operations often introduce substantial computational and memory overhead, limiting their efficiency on large atomic systems such as catalyst slabs. $E^{2}$GNN addresses this issue by adopting a lightweight scalar--vector representation, aiming to balance prediction accuracy, equivariance, and computational efficiency.

\textbf{Scalar--Vector Dual Representation}
Instead of relying on high-order irreducible representations, $E^{2}$GNN represents each atom using both scalar and vector features:
\[
\mathbf{h}_i = \left( \mathbf{s}_i, \mathbf{v}_i \right),
\]
where $\mathbf{s}_i$ denotes rotation-invariant scalar features and $\mathbf{v}_i$ denotes rotation-equivariant vector features. The scalar channel captures chemical and local-environment information, while the vector channel preserves directional geometric information.

\textbf{Equivariant Message Passing}
For each edge $(i,j)$, $E^{2}$GNN constructs messages based on neighboring atomic features and relative position vectors:
\[
\mathbf{r}_{ij} = \mathbf{x}_j - \mathbf{x}_i.
\]
The message-passing process updates scalar and vector features jointly, allowing the model to encode both distance-dependent interactions and directional structural information. A general update form can be written as:
\[
\mathbf{s}'_i = \phi_s \left( \mathbf{s}_i, \sum_{j \in \mathcal{N}(i)} \mathbf{m}^{s}_{ij} \right),
\]
\[
\mathbf{v}'_i = \phi_v \left( \mathbf{v}_i, \sum_{j \in \mathcal{N}(i)} \mathbf{m}^{v}_{ij} \right),
\]
where $\mathbf{m}^{s}_{ij}$ and $\mathbf{m}^{v}_{ij}$ denote scalar and vector messages, respectively. This formulation enables the network to preserve equivariance while avoiding expensive high-order tensor operations.

\textbf{Efficiency-Oriented Design}
The efficiency of $E^{2}$GNN mainly comes from its use of low-order scalar--vector features. Compared with models that employ high-degree spherical harmonics or tensor products, $E^{2}$GNN reduces both memory consumption and computational cost. This makes it suitable for larger-scale atomistic prediction tasks where efficiency is a major constraint.

\textbf{Performance Characteristics}
$E^{2}$GNN is designed to provide accurate and efficient energy and force predictions across molecules and crystals. By explicitly encoding geometric symmetry while maintaining a lightweight architecture, it offers a favorable trade-off between accuracy and speed.

\textbf{Comparative Advantages}
\begin{itemize}
    \item Lower computational cost than high-order equivariant models
    \item Explicit preservation of directional geometric information
    \item Efficient scalar--vector representation for atomistic systems
    \item Suitable for large-scale energy and force prediction tasks
\end{itemize}

\textbf{Remaining Challenges}
\begin{itemize}
    \item Potentially lower expressiveness than high-order equivariant Transformer architectures
    \item Limited ability to model complex many-body angular interactions compared with tensor-product-based methods
    \item Primarily focused on geometric modeling, with limited integration of textual or symbolic domain knowledge
\end{itemize}

\textbf{Role in Catalysis Benchmarks}
In this work, $E^{2}$GNN is included as an efficient equivariant GNN baseline. Its comparison with CatalyticMLLM helps evaluate whether multimodal graph--language modeling can improve adsorption energy prediction beyond lightweight geometric equivariant architectures.

\subsection{DiffCSP}

DiffCSP is a diffusion-based crystal structure prediction and generation method designed for periodic crystalline systems. It learns the distribution of stable crystal structures by jointly generating lattice parameters and atomic fractional coordinates under periodic boundary conditions \cite{jiao2023diffcsp}.

\textbf{Design Philosophy}
DiffCSP aims to adapt diffusion generative modeling to crystal structure prediction, where the generated structures must satisfy translation invariance, rotation invariance, atom permutation invariance, and lattice periodicity. Instead of treating a crystal as a simple molecular graph, DiffCSP explicitly models the periodic geometry of crystals.

\textbf{Crystal Representation}
Each crystal structure is represented by:
\begin{itemize}
    \item Atomic species
    \item Lattice matrix
    \item Fractional atomic coordinates
\end{itemize}
The use of fractional coordinates is important because it naturally conforms to periodic boundary conditions and avoids discontinuities at unit-cell boundaries.

\textbf{Joint Equivariant Diffusion}
DiffCSP performs diffusion over both lattice variables and atomic coordinates. During the forward process, noise is gradually added to the crystal representation. During the reverse process, a periodic $E(3)$-equivariant denoising network reconstructs the crystal structure step by step.

\textbf{Mathematical Formulation}
The denoising model can be written abstractly as:
\[
(\hat{\mathbf{L}}, \hat{\mathbf{X}}) =
\epsilon_{\theta}(\mathbf{L}_{t}, \mathbf{X}_{t}, t, \mathbf{Z}),
\]
where $\mathbf{L}_{t}$ denotes the noisy lattice, $\mathbf{X}_{t}$ denotes noisy fractional coordinates, $\mathbf{Z}$ denotes atom types, and $t$ is the diffusion timestep.

\textbf{Physical Inductive Bias}
DiffCSP incorporates:
\begin{itemize}
    \item Periodic boundary conditions
    \item Rotation and translation equivariance
    \item Joint lattice-coordinate generation
    \item Fractional-coordinate-based position modeling
\end{itemize}

\textbf{Training Objective}
The model is trained to recover clean crystal structures from noisy inputs, typically using denoising losses on lattice parameters and fractional coordinates.

\textbf{Strengths}
\begin{itemize}
    \item Suitable for periodic crystal generation
    \item Jointly generates lattice and atomic positions
    \item Strong geometric inductive bias through periodic equivariance
\end{itemize}

\textbf{Limitations}
\begin{itemize}
    \item Primarily optimized for crystal structure generation rather than catalytic adsorption inverse design
    \item Does not inherently unify structure generation and property prediction
    \item Usually requires an external evaluator for property-conditioned screening
\end{itemize}

\textbf{Role in Benchmarking}
DiffCSP serves as a representative diffusion-based baseline for crystal and catalytic-material structure generation. In inverse design experiments, it reflects the performance of a conventional generative model under a decoupled generation--evaluation paradigm.

\subsection{CrystalFlow}

CrystalFlow is a flow-based generative model for crystalline materials. Unlike diffusion models that generate CIF through iterative denoising, CrystalFlow uses continuous normalizing flows and conditional flow matching to transform a simple prior distribution into the target distribution of crystal structures \cite{luo2025crystalflow}.

\textbf{Design Philosophy}
CrystalFlow is motivated by the need for an efficient and continuous generative process for crystal structures. It formulates crystal generation as learning a transport path from a prior distribution to the data distribution.

\textbf{Crystal Representation}
A crystal is represented by:
\begin{itemize}
    \item Atom types
    \item Lattice parameters
    \item Atomic coordinates
    \item Optional composition or structure-level conditions
\end{itemize}

\textbf{Continuous Normalizing Flow}
CrystalFlow models the probability flow from a simple base distribution $p_0$ to the crystal data distribution $p_1$. The generation process is governed by an ordinary differential equation:
\[
\frac{d\mathbf{x}_{t}}{dt} = v_{\theta}(\mathbf{x}_{t}, t, \mathbf{c}),
\]
where $\mathbf{x}_{t}$ denotes the intermediate crystal representation, $v_{\theta}$ is the learned velocity field, and $\mathbf{c}$ denotes conditional information.

\textbf{Conditional Flow Matching}
Instead of explicitly computing likelihoods at every step, CrystalFlow learns the velocity field by conditional flow matching:
\[
\mathcal{L}_{\mathrm{CFM}} =
\mathbb{E}_{t,\mathbf{x}_{t}}
\left[
\left\|
v_{\theta}(\mathbf{x}_{t},t,\mathbf{c}) - u_{t}
\right\|^{2}
\right],
\]
where $u_t$ is the target velocity along the probability path.

\textbf{Physical Inductive Bias}
CrystalFlow incorporates:
\begin{itemize}
    \item Symmetry-aware crystal representation
    \item Graph-based equivariant neural networks
    \item Continuous and invertible generative dynamics
\end{itemize}

\textbf{Strengths}
\begin{itemize}
    \item Efficient sampling through learned continuous flows
    \item Suitable for conditional crystal generation
    \item Avoids some instability associated with long diffusion chains
\end{itemize}

\textbf{Limitations}
\begin{itemize}
    \item Requires carefully designed flow paths and velocity fields
    \item Property optimization is usually performed through an external scoring model
    \item Does not directly guarantee target relaxed-energy matching for catalytic adsorption systems
\end{itemize}

\textbf{Role in Benchmarking}
CrystalFlow provides a strong flow-based baseline complementary to diffusion-based methods. It is useful for evaluating whether continuous transport-based generation can produce valid catalytic-material candidates under the same inverse-design setting.

\subsection{CrysText}

CrysText is a text-conditioned crystal structure generation framework based on large language models. It directly generates crystal structures in Crystallographic Information File (CIF) format from textual prompts, such as composition, space group, and natural-language structural descriptions \cite{mohanty2025crystext}.

\textbf{Design Philosophy}
CrysText treats crystal generation as a sequence modeling problem. Since CIF files are structured text files containing lattice parameters, symmetry information, and atomic coordinates, an autoregressive language model can learn CIF syntax and structural regularities from large-scale crystallographic data.

\textbf{Input and Output Format}
The input of CrysText is a text prompt describing the desired crystal, while the output is a CIF sequence:
\[
x_{\mathrm{text}} \longrightarrow C_{\mathrm{CIF}}.
\]
The generated CIF can then be parsed by crystallographic toolkits for structure validation and downstream simulation.

\textbf{Autoregressive CIF Generation}
The generation probability is factorized token by token:
\[
p(C_{\mathrm{CIF}} \mid x_{\mathrm{text}})
=
\prod_{t=1}^{T}
p(c_t \mid c_{<t}, x_{\mathrm{text}}),
\]
where $c_t$ is the $t$-th CIF token.

\textbf{Structural Knowledge in Text}
CrysText exploits the fact that CIF files encode:
\begin{itemize}
    \item Unit-cell parameters
    \item Space-group information
    \item Atomic-site labels
    \item Fractional coordinates
    \item Element composition
\end{itemize}

\textbf{Strengths}
\begin{itemize}
    \item Directly generates human-readable CIF files
    \item Naturally supports text-conditioned crystal generation
    \item Compatible with large language model pretraining and fine-tuning
\end{itemize}

\textbf{Limitations}
\begin{itemize}
    \item May generate syntactically invalid CIF files
    \item May suffer from composition mismatch or missing structural fields
    \item Lacks explicit three-dimensional geometric modeling during generation
\end{itemize}

\textbf{Role in Benchmarking}
CrysText is an important LLM-based baseline for CIF-level structure generation. It is especially relevant for comparing against CatalyticMLLM because both methods use textual representations and autoregressive generation, but CatalyticMLLM further incorporates 3D structural information and unified property prediction.

\subsection{CrysText-RL}

CrysText-RL is a reinforcement-learning-enhanced variant of CrysText. It extends the supervised autoregressive CIF generator by introducing reward-based optimization to improve validity, structural realism, and target consistency of generated crystal structures \cite{mohanty2025crystext}.

\textbf{Design Philosophy}
While standard CrysText learns to imitate CIF sequences from training data, CrysText-RL further guides the language model toward high-quality generations using task-specific rewards. This makes the model more suitable for inverse design scenarios where syntactic correctness and structural feasibility are both required.

\textbf{Reward-Guided Generation}
For each text prompt, the model samples multiple CIF candidates:
\[
\{C_1,\ldots,C_K\}\sim \pi_{\theta}(\cdot \mid x_{\mathrm{text}}).
\]
Each candidate is evaluated by a reward function involving structural and text-condition constraints.

\textbf{Typical Reward Components}
The reward may include:
\begin{itemize}
    \item CIF format validity
    \item Parseability
    \item Composition consistency
    \item Structural realism
    \item Similarity to the target condition or target structure
\end{itemize}

\textbf{Policy Optimization}
A reinforcement learning objective is used to increase the probability of high-reward CIF sequences:
\[
\mathcal{L}_{\mathrm{RL}}
=
-
\mathbb{E}_{C\sim\pi_{\theta}}
\left[
R(C)\log \pi_{\theta}(C\mid x_{\mathrm{text}})
\right].
\]
A KL regularization term can be used to prevent the updated policy from drifting too far from the supervised CrysText model.

\textbf{Strengths}
\begin{itemize}
    \item Improves CIF validity compared with pure supervised generation
    \item Can incorporate multiple structural constraints through reward design
    \item Retains the flexibility of language-model-based CIF generation
\end{itemize}

\textbf{Limitations}
\begin{itemize}
    \item Reward design strongly affects final generation quality
    \item Evaluation still relies on external validation or reward functions
    \item Does not fully couple property prediction and structure generation in a shared multimodal representation space
\end{itemize}

\textbf{Role in Benchmarking}
CrysText-RL is used as a reinforcement-learning baseline for CIF generation. It helps evaluate whether the improvement of CatalyticMLLM comes merely from reward tuning or from the unified graph--text multimodal architecture and closed-loop optimization strategy.

\subsection{CatDRX}

CatDRX is a reaction-conditioned generative model for catalyst design and optimization. It is designed to generate catalyst candidates under given reaction conditions by learning the relationship between catalyst structures, reaction components, and catalytic performance \cite{CatDRX}.

\textbf{Design Philosophy}
The main motivation of CatDRX is that catalyst performance is not determined by catalyst structure alone, but also depends strongly on the reaction context. Therefore, CatDRX conditions the generative process on reaction information, enabling reaction-specific catalyst generation.

\textbf{Reaction-Conditioned Representation}
Reaction conditions may include:
\begin{itemize}
    \item Reactants
    \item Products
    \item Reagents
    \item Solvents or additives
    \item Reaction time or other experimental descriptors
\end{itemize}

\textbf{Variational Autoencoder Framework}
CatDRX is built around a conditional variational autoencoder. The encoder maps catalyst and reaction information into a latent variable:
\[
q_{\phi}(\mathbf{z}\mid \mathbf{c}, \mathbf{r}),
\]
where $\mathbf{c}$ denotes catalyst representation and $\mathbf{r}$ denotes reaction condition representation. The decoder generates catalyst candidates conditioned on $\mathbf{z}$ and $\mathbf{r}$:
\[
p_{\theta}(\mathbf{c}\mid \mathbf{z}, \mathbf{r}).
\]

\textbf{Training Objective}
The training objective follows the standard conditional VAE form:
\[
\mathcal{L}_{\mathrm{VAE}}
=
\mathcal{L}_{\mathrm{rec}}
+
\mathrm{KL}
\left[
q_{\phi}(\mathbf{z}\mid \mathbf{c},\mathbf{r})
\|p(\mathbf{z})
\right],
\]
where $\mathcal{L}_{\mathrm{rec}}$ reconstructs catalyst representations and the KL term regularizes the latent space.

\textbf{Catalyst Optimization}
After learning a reaction-conditioned latent space, CatDRX can sample or optimize latent variables to propose catalyst candidates suitable for a given reaction. A performance prediction module may be used to estimate catalytic outcomes such as yield or selectivity.

\textbf{Strengths}
\begin{itemize}
    \item Explicitly incorporates reaction context
    \item Supports reaction-specific catalyst generation
    \item Provides a latent space for catalyst optimization
\end{itemize}

\textbf{Limitations}
\begin{itemize}
    \item Mainly designed for reaction-conditioned molecular or catalyst discovery rather than CIF-level catalytic adsorption structure generation
    \item Often relies on a separated generation and evaluation workflow
    \item Fine-grained relaxed-energy matching is not directly encoded as a unified generation--prediction task
\end{itemize}

\textbf{Role in Benchmarking}
CatDRX is included as a catalyst-oriented generative baseline. Compared with general crystal generation methods, it is more closely related to catalyst discovery, but it still differs from CatalyticMLLM because it does not unify 3D structural encoding, textual prompting, property prediction, and CIF generation in one multimodal model.

\subsection{MAGECS}

MAGECS, short for Material Generation with Efficient Global Chemical Space Search, is an inverse-design framework that combines generative models, supervised graph neural networks, and the bird swarm algorithm for efficient exploration of large chemical spaces \cite{MAGECS}.

\textbf{Design Philosophy}
The central motivation of MAGECS is that conventional generative models often explore only a localized region of the chemical space. To overcome this limitation, MAGECS introduces a population-based global search strategy to guide the generative model toward materials with target properties.

\textbf{Framework Components}
MAGECS consists of three major components:
\begin{itemize}
    \item A generative model for proposing material candidates
    \item A supervised graph neural network for property prediction
    \item A bird swarm algorithm for global chemical-space search
\end{itemize}

\textbf{Property-Guided Search}
The supervised GNN evaluates generated candidates and provides property scores. These scores guide the bird swarm algorithm to update the search direction in chemical or latent space:
\[
\mathbf{z}^{(t+1)}
=
\textsc{BSA}
\left(
\mathbf{z}^{(t)}, f_{\mathrm{GNN}}(\mathbf{z}^{(t)})
\right),
\]
where $\mathbf{z}^{(t)}$ denotes the population state at iteration $t$, and $f_{\mathrm{GNN}}$ denotes the property predictor.

\textbf{Bird Swarm Algorithm}
The bird swarm algorithm is a population-based optimization method. In MAGECS, it helps balance:
\begin{itemize}
    \item Global exploration of broad chemical space
    \item Local exploitation around promising candidates
    \item Property-driven candidate selection
\end{itemize}

\textbf{Application Scenario}
MAGECS has been applied to the inverse design of alloy electrocatalysts for CO$_2$ reduction. By combining generation and global search, it can increase the proportion of candidates satisfying desired catalytic properties.

\textbf{Strengths}
\begin{itemize}
    \item Efficient global chemical-space exploration
    \item Combines generative modeling with property-guided optimization
    \item Suitable for large-scale electrocatalyst screening
\end{itemize}

\textbf{Limitations}
\begin{itemize}
    \item Generator, predictor, and search algorithm are separated components
    \item May suffer from evaluator bias if the predictor is unreliable outside its training distribution
    \item Does not perform unified CIF generation and relaxed-energy prediction within a single model
\end{itemize}

\textbf{Role in Benchmarking}
MAGECS is a strong population-based inverse-design baseline. It is particularly useful for comparing against GA-GRPO because both involve candidate-pool evolution. However, MAGECS follows a decoupled optimization paradigm, whereas GA-GRPO performs generation, evaluation, and policy refinement within the unified CatalyticMLLM framework.

\section{Computational Resources.}
All experiments were conducted on a single server equipped with 8$\times$ NVIDIA A100 GPUs (80GB HBM2e each; 640GB total GPU memory) interconnected via NVLink. 
The server uses dual AMD EPYC 7763 CPUs (128 physical cores in total), 1TB DDR4 system memory, and 2$\times$ 3.84TB NVMe SSDs for local storage.
We run Ubuntu 20.04 with CUDA 11.8 and cuDNN 8.x, and implement our models in PyTorch. 
Mixed-precision training (fp16) is enabled to improve throughput and reduce memory usage.

\end{document}